%% file: example_paper.tex
\documentclass{article}

\usepackage{microtype}
\usepackage{graphicx}
\usepackage{subcaption}
\usepackage{booktabs} 

\usepackage{hyperref}

\usepackage[preprint]{icml2026}

\usepackage{amsmath}
\usepackage{amssymb}
\usepackage{mathtools}
\usepackage{amsthm}

\usepackage{colortbl}
\usepackage{caption}
\usepackage{xcolor}
\usepackage{enumitem}
\usepackage{tabularx}
\usepackage{multirow}
\usepackage{makecell}

\definecolor{mygreen}{RGB}{0,160,0}

\definecolor{myred}{RGB}{178,34,34}

\usepackage[capitalize,noabbrev]{cleveref}

\theoremstyle{plain}

\theoremstyle{definition}

\theoremstyle{remark}

\usepackage[textsize=tiny]{todonotes}

\icmltitlerunning{MetaphorVU: Towards Metaphorical Video Understanding}

\begin{document}

\twocolumn[
  \icmltitle{MetaphorVU: Towards Metaphorical Video Understanding}



  \icmlsetsymbol{equal}{*}

  \begin{icmlauthorlist}
    \icmlauthor{Zhuoqun Li}{iscas,ucas}
    \icmlauthor{Boxi Cao}{iscas}
    \icmlauthor{Guiping Jiang}{kuaishou}
    \icmlauthor{Fangrui Lv}{thu}
    \icmlauthor{Ruotong Pan}{kuaishou}
    \icmlauthor{Jianan Wang}{kuaishou}
    \icmlauthor{Xiangyu Wu}{kuaishou}
    \icmlauthor{Hongyu Lin}{iscas}
    \icmlauthor{Yaojie Lu}{iscas}
    \icmlauthor{Yong Du}{kuaishou}
    \icmlauthor{Ruyin Jia}{kuaishou}
    \icmlauthor{Liyan}{kuaishou}
    \icmlauthor{Tingting Gao}{kuaishou}
    \icmlauthor{Han Li}{kuaishou}
    \icmlauthor{Xianpei Han}{iscas}
    \icmlauthor{Le Sun}{iscas}
  \end{icmlauthorlist}

  \icmlaffiliation{iscas}{Chinese Information Processing Laboratory, Institute of Software, Chinese Academy of Sciences}
  \icmlaffiliation{ucas}{University of Chinese Academy of Sciences}
  \icmlaffiliation{thu}{Department of Automation, Tsinghua University}
  \icmlaffiliation{kuaishou}{Kuaishou Technology}

  \icmlcorrespondingauthor{Boxi Cao}{caoboxi@iscas.ac.cn}
  \icmlcorrespondingauthor{Xiangyu Wu}{wuxiangyu06@kuaishou.com}

  \icmlkeywords{Machine Learning, ICML}

  \vskip 0.3in
]



\printAffiliationsAndNotice{}  

\begin{abstract}

\renewcommand{\thefootnote}{\fnsymbol{footnote}}
\setcounter{footnote}{0}

Metaphorical videos are prevalent across various real-world scenarios to convey complex ideas, and understanding them typically requires high-order cognitive capabilities. 
The lack of systematic studies on metaphorical video understanding not only constrains the real-world applicability of MLLMs but also impedes the thorough assessment of their high-order cognitive capabilities.
To bridge this gap, we propose MetaphorVU-Bench, the first systematic and comprehensive benchmark dedicated to metaphorical video understanding. 
Through experiments, we find current MLLMs struggle with accurate metaphorical video understanding, lagging far behind human level, primarily due to defective cross-domain mapping. 
Motivated by this finding, we construct a metaphor knowledge graph as mapping augmentation and propose MetaphorBoost, an inference-time enhancement framework achieving consistent performance improvement. 
Our benchmark, analysis, and method provide useful insights and a foundation for future research on advancing MLLMs. 
Code: \textcolor{blue}{https://github.com/icip-cas/MetaphorVU}.


\end{abstract}

\input{sections/introduction5}

\input{sections/benchmark}

\input{sections/experiment}

\input{sections/method}

\input{sections/related}

\input{sections/conclusion}

\section*{Impact Statement}
This paper presents work whose goal is to advance the field of machine learning. There are many potential societal consequences of our work, none of which we feel must be specifically highlighted here.

\section*{Acknowledgments}
We sincerely thank the reviewers for their insightful comments and valuable suggestions. This work was supported by the National Key R\&D Program of China (2024YFC3308000), the Natural Science Foundation of China (No. 62536008, 62476265, 62306303).


\bibliography{example_paper}
\bibliographystyle{icml2026}

\newpage
\appendix
\onecolumn

\input{apps/theory}
\input{apps/construction}

\input{apps/evaluation}

\input{apps/extract_and_datasets}

\input{apps/identify_and_generate}
\input{apps/baselines}

\input{apps/hyperpara}
\input{apps/case}

\newpage
\input{figs/case_study}
\input{figs/prompt_for_LLM_fil}
\input{figs/prompt_for_MLLM_fil}
\input{figs/prompt_for_human_fil}
\input{figs/prompt_for_human_ann}
\input{figs/prompt_for_evalu}
\input{figs/prompt_for_judge}
\input{figs/prompt_for_extract_pair}
\input{figs/prompt_for_identi}
\input{figs/prompt_for_gene}
\input{figs/more_case_1}
\input{figs/more_case_2}
\input{figs/more_case_3}
\input{figs/more_case_4}
\input{figs/more_case_5}
\input{figs/more_case_6}
\input{figs/more_case_7}
\input{figs/more_case_8}


\end{document}

%% file: sections/introduction5.tex
\section{Introduction}

Metaphorical videos serve as a crucial medium for conveying complex ideas in human society, and they widely exist in important scenarios such as social media and public communication~\cite{krippendorff1993major,shifman2013memes,burgers2016figurative,shutsko2020user}.
Rather than directly presenting profound meanings such as society criticism and life contemplation, video creators  often employ metaphorical content to guide viewers toward  associations and interpretations~\cite{johnson1979some,camac1984metaphors,zhang2021visual,alnajjar2022ring}.
According to multimodal metaphor theory, human understanding of metaphorical videos is a high-order cognitive process that transforms perceived signals into deeper semantics, with the core lying in cross-domain mapping that links visual elements to underlying concepts~\cite{forceville2009non,fahlenbrach2016embodied,pan2020identifying,zhang2021visual}.
As illustrated in Figure~\ref{fig:head}, humans can link visual elements (e.g., \textit{tailcoat pigs, banquet, and cats under table}) with underlying concepts (e.g., \textit{ruling group, social wealth, and underprivileged}), thereby revealing implicit meanings of \textit{critique toward the ruling group and sympathy for the lower class people}.

\input{figs/head}

Recently, multimodal large language models (MLLMs) have been widely used in practical applications and  significantly pushed the frontier of video understanding capabilities~\cite{gpt5,bai2025qwen3vltechnicalreport,an2025llava,gmini3}. 
Unfortunately, most existing work focuses on literal perception tasks such as object recognition and event description of videos~\cite{li2025survey,bandraupalli2025vlms,brkic2025frame,liu2025surveillancevqa}, lacking a systematic study of high-order cognitive metaphorical video understanding.
This gap makes it difficult to assess whether MLLMs can  accurately transform  perceived visual signals into deeper semantics like humans, limiting their reliable application in many complex scenarios and further improvement of cognitive capabilities~\cite{shutsko2020user,zhang2021visual,alnajjar2022ring,okonski2022understanding}.
Therefore, effectively evaluating and advancing the metaphorical video understanding capability of MLLMs is of great significance for their widespread utilization and further enhancement.

To this end, we propose \textbf{MetaphorVU-Bench}\footnote{The proposed benchmark of this paper is released in \textcolor{blue}{https://huggingface.co/datasets/lzq2021/MetaphorVU-Bench}.}, 
the first comprehensive benchmark for metaphorical video understanding, characterized by a well-founded systematic taxonomy, metaphorical videos curated from billions of real-world candidates, and rigorous human annotation.
Specially, to ensure a systematic evaluation, as illustrated in Figure~\ref{fig:bench}, we first design a well-founded video metaphor taxonomy, covering 8 types of video metaphor grounded in multimodal metaphor theory~\cite{forceville2009non,forceville2009multimodal} and its extensions~\cite{bordwell2013viewer,stam2017film,schechner2017performance,chandler2022semiotics}. 
Guided by this taxonomy, as illustrated in Figure~\ref{fig:construct}, we construct the benchmark sourced from the real world with careful filtration and rigorous annotation.
Firstly, to ensure the evaluation accurately reflects practical performance, we source data from a real-world video platform covering diverse topics.
Secondly, to efficiently select metaphorical videos from billions of sources, we apply a multi-stage filtration based on video information and  comments, yielding 860 videos spanning the taxonomy.
Finally, to obtain reliable metaphor interpretations, we conduct manual annotation with strict cross-validation, yielding a high-quality benchmark for systematic evaluation of metaphorical video understanding.

Based on above MetaphorVU-Bench, we systematically evaluate 11 representative close-source and open-source MLLMs.
Experimental results show that current MLLMs still struggle with accurate metaphorical video understanding. 
Even the most advanced MLLMs, such as Gemini-3-Pro and GPT-5, can only achieve average scores around 64, significantly lagging behind human-level performance by nearly 20 points.
Furthermore, to better understand causes of MLLM failures and develop targeted optimization methods, we conduct an error analysis across MLLMs of varying capabilities.
Analysis results reveal that over 80\% of failures do not stem from recognition error, but rather from defective cross-domain mapping, where current MLLMs fail to effectively establish links from visual elements to underlying concepts.
These findings indicate that enhancing cross-domain mapping is the key to improving MLLMs performance on metaphorical video understanding.

Motivated by above findings, rather than relying on MLLMs to perform blind cross-domain mapping, we propose a novel enhancing framework, \textbf{MetaphorBoost}, utilizing a metaphorical knowledge graph as external cognitive scaffold to augment cross-domain mapping. 
Specifically, to provide MLLMs with metaphor-specific interconnected augmentation, we construct the first metaphorical knowledge graph by collecting metaphorical texts, extracting metaphorical concepts and connecting these concepts.
At inference time, MetaphorBoost queries the metaphorical knowledge graph based on content recognition results to obtain reliable references, thereby promoting cross-domain mapping and precise metaphor interpretations.
Experimental results show MetaphorBoost achieves consistent performance improvements across multiple MLLMs, providing a preliminary exploration and foundation for future research.
Main contributions of this paper can be summarized as follows:

\vspace{-0.5\baselineskip}
\begin{itemize}
\setlength{\topsep}{0pt}
\setlength{\parskip}{0pt}
\setlength{\itemsep}{2pt}
\setlength{\parsep}{0pt}
\item We propose MetaphorVU-Bench, which is the first benchmark dedicated to systematic and comprehensive evaluation for metaphorical video understanding.
\item We conduct extensive experiments and analysis, revealing  the deficiencies of current MLLMs and providing insights into the underlying causes of their failures.
\item We construct MetaphorBoost, boosting metaphorical video understanding via inference-time mapping augmentation based on a metaphorical knowledge graph.
\end{itemize}
\vspace{-0.5\baselineskip}

%% file: figs/head.tex
\begin{figure}[t!]
\centering
\includegraphics[width=\linewidth]{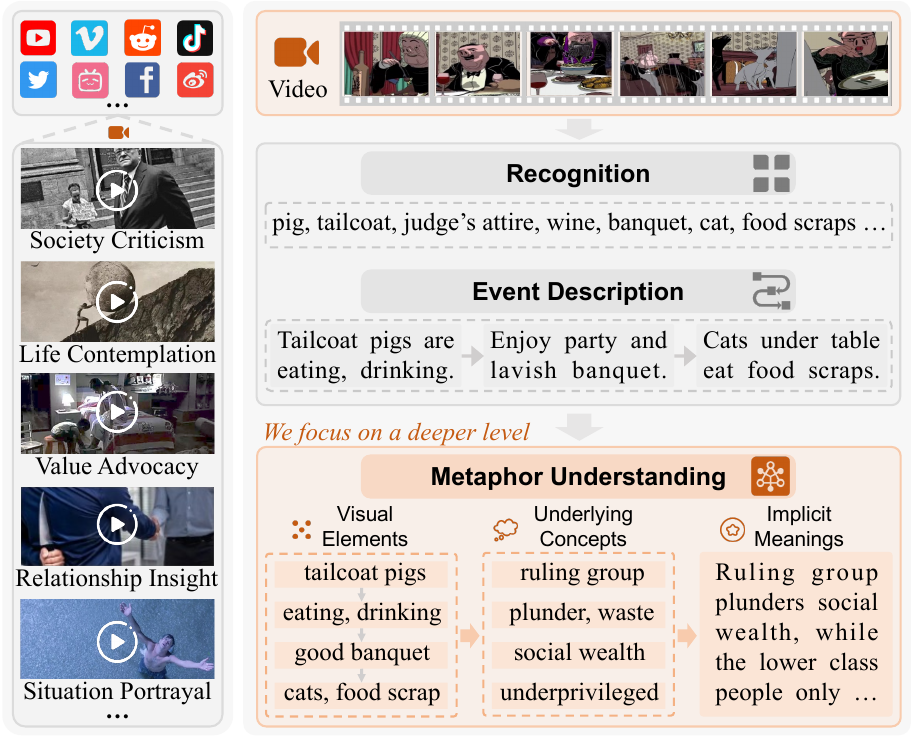} 
\caption{Metaphorical videos are prevalent across various
real-world scenarios to convey many complex ideas, and metaphorical video understanding requires high-order cognitive capabilities.}
\label{fig:head}
\end{figure}


%% file: sections/benchmark.tex
\section{MetaphorVU-Bench}

\input{figs/bench}

The lack of systematic research on metaphorical video understanding to some extent limits further application reliability and capability enhancement of MLLMs.
To bridge this gap, we design the first systematic video metaphor taxonomy and construct MetaphorVU-Bench based on this taxonomy, enabling systematic evaluation of metaphorical video understanding.
In this section, we sequentially present the taxonomy, benchmark and evaluation method.

\subsection{Video Metaphor Taxonomy}
To ensure reliable and principled evaluation of metaphorical video understanding, a systematic video metaphor taxonomy is essential for building the benchmark. 
Therefore, we draw on multimodal metaphor theory~\cite{forceville2009non,forceville2009multimodal} and its extensions in the video field~\cite{bordwell2013viewer,stam2017film,schechner2017performance,chandler2022semiotics}, designing the first systematic video metaphor taxonomy.
Specifically, as illustrated in Figure~\ref{fig:bench}, video metaphor can be categorized as following 8 types: 
\vspace{-0.5\baselineskip}
\begin{itemize}
\setlength{\topsep}{0pt}
\setlength{\parskip}{0pt}
\setlength{\itemsep}{2pt}
\setlength{\parsep}{0pt}
\item \textit{Body Language.} Video conveys implicit meanings through body movements of characters, typically some exaggerated or semantically meaningful actions.
\item \textit{Atmosphere Language.} Video conveys implicit meanings by environmental atmosphere, such as purposeful variations in the color, lighting and composition.
\item \textit{Cultural Symbol.} Video conveys implicit meanings by symbolism of cultural artifacts, such as flying China Kongming lanterns or building a Christianity cross.
\item \textit{Naturalistic Symbol.} Video conveys implicit meanings by symbolism of natural elements, such as animal behaviors, plant growth, and changing starry skies.
\item \textit{Causal Montage.} Video conveys implicit meanings through juxtaposing cause-and-effect shots to guide  audiences to infer some causal logic in their brain.
\item \textit{Analogical Montage.} Video conveys implicit meanings by juxtaposing visually or thematically similar shots to guide audiences to infer analogical logic in brain.
\item \textit{Surreal Narrative.} Video conveys implicit meanings through characters and plots transcending physical constraints, such as cartoons and AI-generated videos.
\item \textit{Performative Narrative.} Video conveys  implicit meanings through dramatized storytelling performed by human actors, such as short play in video platforms.
\end{itemize}
\vspace{-0.5\baselineskip}
This video metaphor taxonomy provides a solid foundation for building a comprehensive benchmark and conducting systematic evaluation.
Examples for each type are illustrated in Figure~\ref{fig:bench}. Detailed theoretical basis for the taxonomy is shown in Appendix~\ref{app:theory}, more examples are in Appendix~\ref{app:case}.

\input{figs/construct}

\subsection{Benchmark Construction}

Based on above video metaphor taxonomy, we construct MetaphorVU-Bench, enabling systematic evaluation of metaphorical video understanding.
Specifically, as shown in Figure~\ref{fig:construct}, we select real-world data source, apply efficient multi-stage filtration and perform reliable manual annotation, obtaining the benchmark  with strict quality validation. 
This benchmark encompasses diverse video topics, with sufficient data volume and suitable video duration for evaluation.
Thematic diversity is shown in Figure~\ref{fig:topic}. Statistics of sample number, video duration and token number of golden interpretation are shown in Table~\ref{tab:type_num}. 
In the following, we provide detailed process of  benchmark construction.

\textbf{Real-world Data Source.}
We prioritize diversity and authenticity when selecting data source, which are two critical factors for credible evaluation.
Specially, to ensure evaluation results can accurately reflect metaphorical video understanding capability in real world, the benchmark should cover diverse video topics from daily life. 
Moreover, since current MLLMs mainly support inputting a limited number of frames, the benchmark should contain videos with compatible durations to avoid video length becoming a confounding factor.
Therefore, we use Kuaishou\footnote{https://www.kuaishou.com/?isHome=1} short-video platform as the data source, which can provide massive real-world videos spanning a wide range of topics and video duration is compatible with most common-used MLLMs.

\textbf{Efficient Multi-stage Filtration.}
The data source contains billions of videos, of which only a small fraction involve metaphorical logic. To efficiently isolate metaphorical videos, we design a multi-stage filtration strategy.

Considering audience comments often contain interpretation of videos, which can serve as an important indicator, we first filter videos by amount of audience comments, retaining only those with more than 150 comments, yielding 70K videos.
Then, we use a powerful LLM (GPT-5) to analyze the video introduction, automatic speech recognition (ASR) result and audience comments to determine whether each video contains metaphorical logic, reducing the amount of candidate video set to 16K. The detailed prompt guideline for LLM to do filtration is shown in Appendix~\ref{app:filter_llm}.

Furthermore, considering above filtration process does not directly use visual information and LLM analysis may not align with the actual video, we conduct further check and filtration. 
A powerful MLLM (Gemini-3-Pro) is used to verify whether above analysis is consistent with original videos, reducing the amount of candidate video set to 4K. 
Then, a human team performs final filtration based on original video, video introduction and audience comments, resulting in 860 videos with definite metaphorical logic. 
Additionally, annotators identify the metaphor type for each video, balancing the number of samples across each metaphor type as much as possible.
The prompt for MLLM and human annotators  filtration are in the Appendix~\ref{app:filter_mllm} and \ref{app:filter_human}, respectively.

\input{tabs/bench}
\input{figs/topic}

\textbf{Reliable Manual Annotation.}
Since video metaphor interpretation is a flexible text, different annotators may produce varying linguistic styles and formats. Although these interpretations may all be substantively correct, such subjectivity and format inconsistency make it difficult to conduct evaluation by the benchmark.
Therefore, when annotating video metaphor interpretation, we require human annotators to reference video introduction and audience comments and follow a fixed format (i.e., \textit{specifying which visual elements convey which implicit meanings}). This can reduce subjectivity and enhance format consistency, thereby improving the reliability of benchmark.
Additionally, annotators are responsible for providing a brief title that introduces necessary background information of the video.
The guideline for manual annotation is shown in Appendix~\ref{app:annotation}.

\input{tabs/main}

\textbf{Strict Quality Control.}
To further ensure benchmark quality, we employ cross-validation among annotators to avoid errors by individual oversight.
During the final video filtration stage, we assign three annotators for each candidate video. If any annotator considers the video to lack  definite metaphorical logic, the video is excluded.
During the interpretation annotation stage, we assign one interpreter and two reviewers for each video. The initial annotation from interpreter is reviewed by reviewers, and all three iteratively refine it until reaching a good metaphor interpretation that is acceptable to all.
In additional, to avoid speech and subtitles in videos directly unveiling the metaphorical meanings, we apply muting and subtitle removal using open-source tool\footnote{https://github.com/YaoFANGUK/video-subtitle-remover} before manual annotation, ensuring both annotation and evaluation rely solely on visual information of videos.

\subsection{Evaluation Task and Metric} 
\textbf{Task Formulating.}
Based on this benchmark, we evaluate the metaphorical video understanding as following formula:
\begin{equation}
\hat{\tau}, \hat{o} = \mathcal{F}(v \oplus t)
\end{equation} 
where $\mathcal{F}$ is evaluated system, $v$ is video,  $t$ is title, $\oplus$ denotes input combination, $\hat{\tau}$ is thinking process and $\hat{o}$ is output video metaphor interpretation.
Generally, MLLMs first recognize visual elements, establish linking to underlying concepts and reveal implicit meanings in $\hat{\tau}$, then formally interpret which visual elements convey which implicit meanings in $\hat{o}$. Detailed evaluation prompt is shown in Appendix~\ref{app:evaluation}.

\textbf{Evaluation Metric.} Since video metaphor interpretation is free-form text, rule-based metrics are difficult to provide reliable scores~\cite{mayfield2024evaluation,li2025benchmark}. 
Therefore, we follow the metrics in previous free-form video-QA works~\cite{yu2025vrbench,long2025adsqa}, using DeepSeek-V3.2\footnote{https://api-docs.deepseek.com/news/news251201} as LLM judge.
Specifically, we design detailed scoring guidelines for LLM judge to accurately assess MLLMs output. With golden interpretation as reference, the judge evaluates output interpretation on its accuracy in grounding metaphorical visual elements and revealing implicit meanings, assigning a integer score from 0 to 10, then rescaled to  0-100 for presentation.
Guidelines for LLM judge are in Appendix~\ref{app:judge}. 
Consistency analysis between LLM judge and human judge is in Appendix~\ref{app:consistency}, where Pearson correlation coefficient is 0.85, confirming the LLM judge is reliable.

%% file: figs/bench.tex
\begin{figure*}[t!]
\centering
\includegraphics[width=\linewidth]{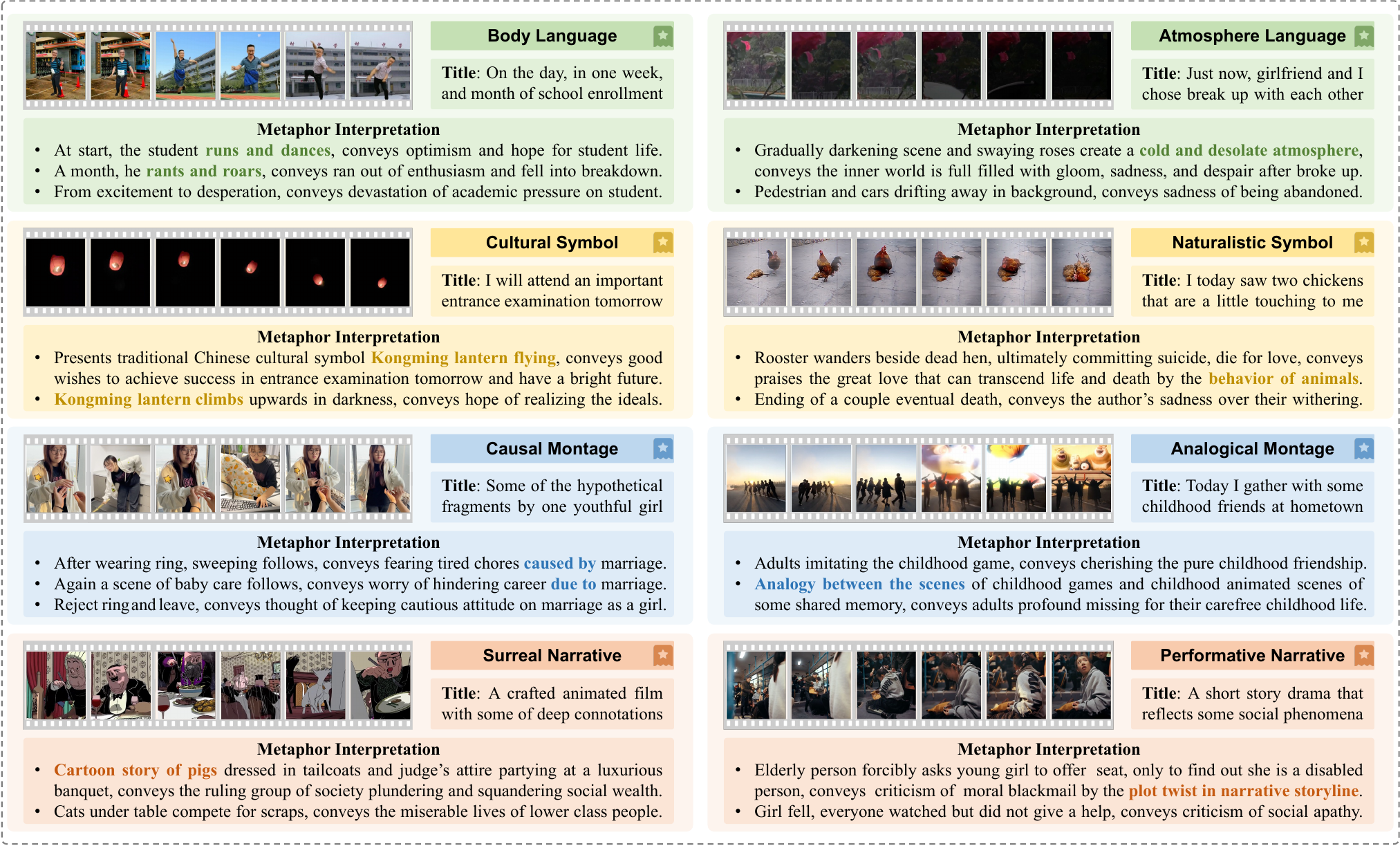} 
\caption{MetaphorVU-Bench contains 8 types of video metaphor, enabling systematic evaluation of metaphorical video understanding. Note that most videos simultaneously contain multiple types of metaphor, we only show the dominant one in each case for illustration.}  
\label{fig:bench}
\end{figure*}

%% file: figs/construct.tex
\begin{figure*}[t!]
\centering
\includegraphics[width=\linewidth]{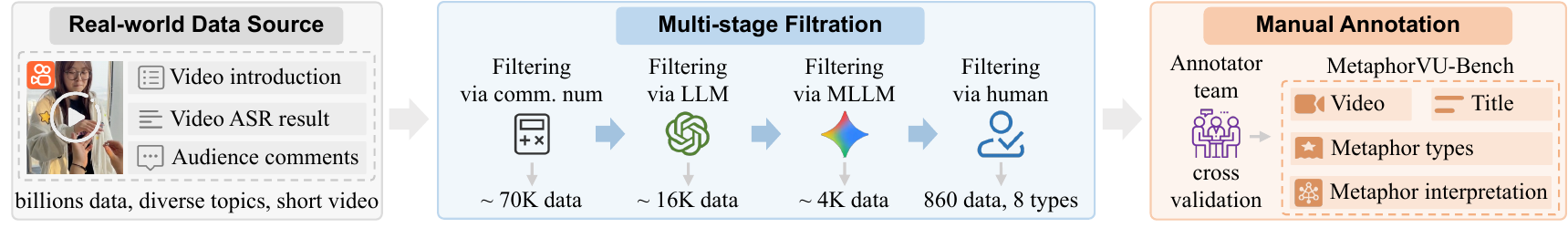} 
\caption{We construct MetaphorVU-Bench by using a real-world short-video platform as source, selecting metaphorical videos from a large-scale video pool through multi-stage filtration, and manually annotating video metaphor interpretations with rigorous quality control. MetaphorVU-Bench can effectively support systematic and comprehensive evaluation of metaphorical video understanding.}  
\label{fig:construct}
\end{figure*}

%% file: tabs/bench.tex


\begin{table}[!tp]
\caption{Benchmark statistics of sample number, average video duration and average token number of golden interpretations.}
\centering
\resizebox{\linewidth}{!}{
\begin{tabular}{lccc}
\toprule
\textbf{Type}                       & \textbf{\# Samples}               & \textbf{Avg. Duration} (s) & \textbf{Avg. Tokens} \\
\midrule
Body Language (Body L.)          & 136    & 32.2   & 111.3    \\
Atmosphere Language (Atmosp. L.)    & 150    & 13.1   & 104.5     \\
Cultural Symbol (Cultural S.)      & 62     & 23.5     & 114.4   \\
Naturalistic Symbol (Natural. S.)     & 113    & 17.3  & 108.8      \\
Causal Montage (Causal M.)    & 54     & 57.7     & 108.9   \\
Analogical Montage (Analog. M.) & 171    & 58.7   & 124.8     \\
Surreal Narrative (Surreal N.)    & 112    & 30.4     & 117.1   \\
Performative Narrative (Perform. N.)    & 62     & 86.8    & 118.6    \\
\midrule
MetaphorVU-Bench       & 860    & 37.2  & 114.2  \\
\bottomrule
\end{tabular}%
}
\label{tab:type_num}
\end{table}%

%% file: figs/topic.tex
\begin{figure}[t!]
\centering
\includegraphics[width=\linewidth]{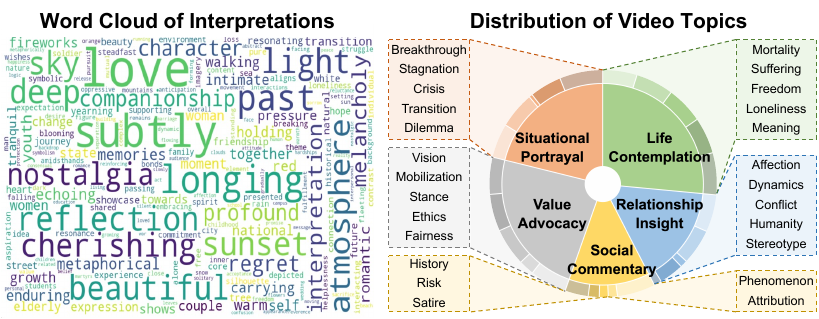} 
\caption{Benchmark covers diverse video topics, enabling  accurate evaluation of real-world metaphorical video understanding.}
\label{fig:topic}
\end{figure}



%% file: tabs/main.tex
\begin{table*}[t!]
\caption{Overall results on MetaphorVU-Bench. To intuitively demonstrate gap between MLLMs and human*, we sample 100 instances and collect human-written metaphor interpretations as  upper-bound. The table shows that current MLLMs exhibit limited capability, and existing reasoning-enhanced methods fail to achieve effective improvements. In contrast, our method proves to be more effective. }
\centering
\resizebox{\linewidth}{!}{
\begin{tabular}{lccccccccc}
\toprule

\textbf{Method} & \textbf{Body L.} & \textbf{Atmosph. L.} & \textbf{Cultural S.} & \textbf{Natural. S.} & \textbf{Causal M.} & \textbf{Analog. M.} & \textbf{Surreal N.} & \textbf{Perform. N.} & \textbf{Average} \\

\midrule
\rowcolor[rgb]{ .906,  .902,  .902} \multicolumn{10}{c}{Upper-bound} \\
\midrule

Human*                            & 87.8 & 87.5 & 89.1 & 83.8 & 72.0 & 81.5 & 78.1 & 78.0 & 83.4 \\

\midrule
\rowcolor[rgb]{ .906,  .902,  .902} \multicolumn{10}{c}{Close-source MLLMs} \\

\midrule
GPT-5~\cite{gpt5}                  & 69.9 & \textbf{76.3} & 77.4 & 66.6 & 45.0 & 55.4 & 54.9 & 46.1 & 63.7 \\
GPT-4o~\cite{hurst2024gpt}                 & 63.4 & 70.5 & 70.3 & 62.6 & 39.1 & 48.2 & 45.7 & 37.9 & 56.8 \\
Qwen3-VL-Plus~\cite{bai2025qwen3vltechnicalreport}          & 66.8 & 72.5 & 74.8 & 65.5 & 51.5 & 54.2 & 50.4 & 43.7 & 61.4 \\
Gemini-2.5-Pro~\cite{gmini2.5}        & 65.5 & 71.3 & 74.3 & 64.4 & 53.5 & 55.7 & 52.1 & 46.9 & 61.8 \\
Gemini-3-Pro~\cite{gmini3}   & 71.2 & 74.0 & 75.1 & \textbf{66.9} & 49.4 & 58.9 & 51.1 & 48.1 & 63.8 \\
Doubao-1.5-Vision-Pro~\cite{guo2025seed1} & 58.2 & 64.1 & 65.5 & 58.9 & 27.8 & 42.5 & 39.8 & 26.6 & 50.5 \\

\midrule
\rowcolor[rgb]{ .906,  .902,  .902} \multicolumn{10}{c}{Open-source MLLMs} \\
\midrule

Qwen2.5-VL-7B-Instruct~\cite{bai2025qwen25vltechnicalreport}      & 36.0 & 49.9 & 46.1 & 42.1 & 12.4 & 23.5 & 28.6 & 16.1 & 33.8 \\
Qwen3-VL-8B-Thinking~\cite{bai2025qwen3vltechnicalreport}        & 56.0 & 66.1 & 68.8 & 60.8 & 33.2 & 45.0 & 39.3 & 29.2 & 52.0 \\
LLaVA-onevision-1.5-8B-Instruct~\cite{an2025llava} & 35.7 & 47.2 & 47.3 & 45.0 & 13.8 & 21.3 & 27.0 & 21.2 & 38.1 \\
GLM-4.5V~\cite{vteam2025glm45vglm41vthinkingversatilemultimodal}                   & 62.7 & 67.9 & 71.9 & 62.1 & 37.6 & 50.1 & 46.1 & 38.4 & 56.8 \\
Qwen3-VL-235B-A22B-Thinking~\cite{bai2025qwen3vltechnicalreport} & 65.4 & 70.4 & 71.9 & 58.1 & 43.2 & 54.6 & 46.1 & 38.1 & 58.6
\\

\midrule
\rowcolor[rgb]{ .906,  .902,  .902} \multicolumn{10}{c}{Reasoning-enhanced Methods} \\
\midrule

VideoRFT~\cite{wang2025videorft}      & 38.9 & 52.8 & 48.4 & 46.0 & 13.5 & 24.8 & 27.2 & 16.6 & 35.6 \\
Vision-R1~\cite{huang2025vision}    & 39.3 & 45.1 & 42.0 & 42.4 & 19.4 & 23.2 & 25.0 & 18.6 & 33.1 \\
ReAd-R~\cite{long2025adsqa}       & 42.1 & 54.1 & 48.9 & 46.3 & 15.7 & 26.4 & 26.2 & 17.6 & 36.8 \\
LTR~\cite{liaodivide}            & 54.1 & 44.7 & 56.2 & 47.4 & 27.8 & 44.6 & 31.9 & 36.1 & 44.5 \\
ViTCoT~\cite{zhang2025vitcot}         & 58.8 & 47.7 & 59.2 & 48.7 & 26.1 & 45.1 & 34.0 & 32.1 & 46.2 \\
Prompt Engineering~\cite{wei2022chain}        & 57.8 & 66.3 & 67.9 & 59.2 & 36.1 & 42.7 & 41.6 & 32.6 & 52.4 \\
Few-shot Example~\cite{dong2024survey}       & 57.6 & 69.4 & 69.2 & 58.7 & 33.5 & 44.9 & 43.5 & 32.6 & 53.6 \\

\midrule
\rowcolor[rgb]{ .906,  .902,  .902} \multicolumn{10}{c}{Mapping Augmentation via Metaphorical Knowledge Graph} \\
\midrule

\rowcolor{blue!5} MetaphorBoost (Gemini-3-Pro) (Ours)            & \textbf{71.5} & \textbf{76.3} & \textbf{77.5} & \textbf{66.9} & \textbf{57.2} & \textbf{59.1} & \textbf{57.3} & \textbf{50.8} & \textbf{66.1} \\
~~$\Delta$ (vs Gemini-3-Pro)             & \textcolor{teal}{+0.3}  & \textcolor{teal}{+2.3}  & \textcolor{teal}{+2.4}  & +0.0  & \textcolor{teal}{+7.8}  & \textcolor{teal}{+0.2}  & \textcolor{teal}{+6.2}  & \textcolor{teal}{+2.8}  & \textcolor{teal}{+2.3}  \\
\rowcolor{blue!5} MetaphorBoost   (Qwen2.5-VL-7B-Instruct) (Ours) & 40.7 & 55.7 & 51.2 & 49.0 & 12.5 & 26.1 & 31.4 & 19.2 & 37.9 \\
~~$\Delta$ (vs Qwen2.5-VL-7B-Instruct)    & \textcolor{teal}{+4.6}  & \textcolor{teal}{+5.8} & \textcolor{teal}{+5.1}  & \textcolor{teal}{+6.9}  & \textcolor{teal}{+0.1}  & \textcolor{teal}{+2.6}  & \textcolor{teal}{+2.9}  & \textcolor{teal}{+3.0}  & \textcolor{teal}{+4.1}  \\
\rowcolor{blue!5} MetaphorBoost   (Qwen3-VL-8B-Thinking) (Ours)   & 61.8 & 71.0 & 71.8 & 61.3 & 36.7 & 47.1 & 45.7 & 31.5 & 55.9 \\
~~$\Delta$ (vs Qwen3-VL-8B-Thinking)      & \textcolor{teal}{+5.8}  & \textcolor{teal}{+4.9}  & \textcolor{teal}{+3.0}  & \textcolor{teal}{+0.5}  & \textcolor{teal}{+3.5}  & \textcolor{teal}{+2.1}  & \textcolor{teal}{+6.4}  & \textcolor{teal}{+2.3}  & \textcolor{teal}{+3.8} \\

\bottomrule
\end{tabular}%
}
\label{tab:main_results}%
\end{table*}%

%% file: sections/experiment.tex
\section{MetaphorVU Evaluation}

\subsection{Evaluation Settings}

\textbf{Selected Baselines.}
To comprehensively evaluate the ability on metaphorical video understanding, we extensively select both close-source and open-source models of various scales, as well as representative reasoning-enhanced methods.
Specially, 
(1) \textbf{Close-source MLLMs},
including GPT-5~\cite{gpt5}, GPT-4o~\cite{hurst2024gpt}, Qwen3-VL-Plus~\cite{bai2025qwen3vltechnicalreport}, Gimini-2.5-Pro~\cite{gmini2.5}, Gimini-3-Pro~\cite{gmini3} and Doubao-1.5-Vision-Pro~\cite{guo2025seed1}. 
(2) \textbf{Open-source MLLMs}, 
including Qwen2.5-VL-7B-Instruct~\cite{bai2025qwen25vltechnicalreport}, Qwen3-VL-8B-Thinking~\cite{bai2025qwen3vltechnicalreport}, LLaVA-onevision-1.5-8B~\cite{an2025llava},  GLM-4.5V~\cite{vteam2025glm45vglm41vthinkingversatilemultimodal}, and the Qwen3-VL-235B-A22B-Thinking~\cite{bai2025qwen3vltechnicalreport}. (3) \textbf{Reasoning-enhanced Methods}, which enhance the reasoning ability of base model by post-training or inference-time scaling, including VideoRFT~\cite{wang2025videorft}, Vision-R1~\cite{huang2025vision}, ReAd-R~\cite{long2025adsqa}, LTR~\cite{liaodivide}, ViTCoT~\cite{zhang2025vitcot},
the first 3 methods are post-training based on Qwen2.5-VL-Instruct, and the last 2 methods are inference-time scaling based on Qwen3-VL-8B-Thinking.
Additionally, we add two commonly used inference-time scaling methods based on Qwen3-VL-8B-Thinking, including
Prompt Engineering~\cite{wei2022chain} with a prompt tailored for metaphorical video understanding, and Few-shot Example~\cite{dong2024survey} with 3-shot examples tailored for metaphorical video understanding. More details of baselines are in Appendix~\ref{app:baselines}.

\textbf{Implementation Details.}
To ensure evaluation reliability, we conduct experiments following the general practices.
For close-source MLLMs, we directly use official APIs for experiments. For open-sourced MLLMs, we download the weights of models from official repositories and deploy them as APIs using vLLM\footnote{https://pypi.org/project/vllm/}. For reasoning-enhanced methods, we use officially provided post-training weights or the inference-time scaling strategies specified in their original papers.
To ensure consistency, the generation temperature is uniformly set to 0.7 for all models.
Regarding the input, since not all MLLMs support direct video input, we follow the common practice by splitting videos into frames and converting them to base64 encoding~\cite{bai2025qwen25vltechnicalreport,bai2025qwen3vltechnicalreport}, thereby supporting all MLLMs involved in this experiment.

\subsection{Overall Results}

Experimental results of MLLMs and reasoning-enhanced methods are in the Table~\ref{tab:main_results}, there are two main conclusions:

\textbf{Current MLLMs struggle
with accurate metaphorical video understanding.}
For open-source MLLMs, table shows there is a significant gap with human, for example, Qwen3-VL-8B-Thinking achieves average score of 52.0, far below the human score of 83.4.
For close-source MLLMs, they can generally achieve relatively higher performance, especially Gemini-3-Pro, demonstrating the strongest overall performance among all baselines, with average  score of 63.8. However, this performance still falls short of the human level, indicating substantial room for improvement.

\textbf{Previous inference-time scaling methods for recognition and event description yield marginal improvement.}
LTR and ViTCoT, which are two inference-time scaling methods designed for enhancing object recognition and event description, even degrade performance of base model Qwen3-VL-8B-Thinking.
In comparison, our implemented prompt engineering and few-shot examples methods designed for metaphorical understanding  yield relatively limited improvements.
Furthermore, despite additional data and training overhead, post-training via long chain-of-thought reinforcement learning  optimized for recognition and description, such as VideoRFT and Vision-R1, only achieve marginal improvements over base model Qwen2.5-VL-Instruct.

\input{tabs/motivation}
\input{figs/different}

\subsection{Detailed Analysis}

\textbf{Error Analysis.}
To  investigate the core deficiencies of MLLMs in detail, we manually observe and identify 4 common types of deficiency in MLLMs thinking process: (1) wrong recognition of visual elements, (2) missing mapping from visual elements to underlying concepts, (3) only superficial mapping,  and (4) improper mapping.
As shown in Appendix Figure~\ref{fig:case_study}, these deficiencies collectively lead to poor output.
Furthermore, to enable more in-depth analysis through quantitative data, we count proportion of each deficiency type.
As shown in Table~\ref{tab:error_p}, incorrect recognition accounts for a small proportion, while  majority is missing, superficial and improper cross-domain mapping.
Therefore, \textit{improving process of linking visual elements to underlying concepts is the key to improving MLLMs performance}.

\textbf{Variations across Metaphor Types.} 
Moreover, we compare MLLMs performance among different video metaphor types. As shown in Figure~\ref{fig:different}, both close-source and open-sourced MLLMs exhibit significantly lower performance on the latter four  types of video metaphor. 
Generally, videos of the latter four types contain richer metaphorical visual elements, whereas the former four types are relatively simpler. Therefore, \textit{MLLMs perform worse on metaphor types requiring more cross-domain mapping, indirectly supporting that mapping augmentation  is the core of improvement}.

%% file: tabs/motivation.tex

\begin{table}[t]
\caption{Proportion of each deficiency type, reveals that enhancing cross-domain mapping is key to improving performance.}
\centering
\resizebox{\linewidth}{!}{
\begin{tabular}{lcccc}
\toprule
\raisebox{0.8ex}{\textbf{Model}} & \textbf{\shortstack[c]{Wrong \\ Recognition}} & \textbf{\shortstack[c]{Missing \\ Mapping}} & \textbf{\shortstack[c]{Superficial \\ Mapping}} & \textbf{\shortstack[c]{Improper \\ Mapping}}\\
\midrule
Gemini-3-Pro & 10.7\% & 27.9\% & 33.7\%  & 27.7\%      \\
Qwen3-VL-8B-Thinking & 13.5\% & 28.1\% & 28.3\% & 30.1\% \\
\bottomrule
\end{tabular}%
\label{tab:error_p}
}
\end{table}%

%% file: figs/different.tex
\begin{figure}[t!]
\centering
\includegraphics[width=\linewidth]{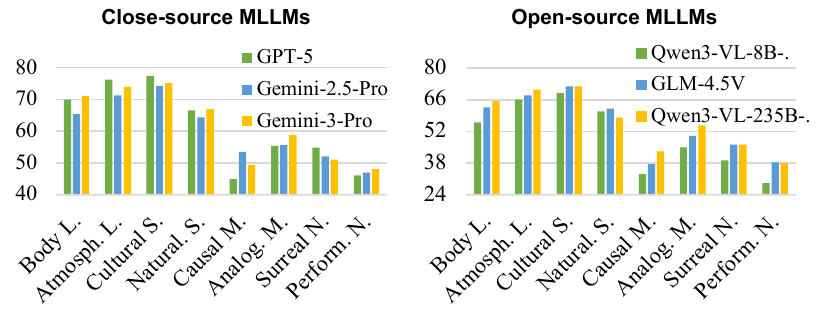} 
\caption{Performing worse on subsets requiring more cross-domain mapping, supports importance of mapping augmentation.}
\label{fig:different}
\end{figure}

%% file: sections/method.tex
\section{MetaphorBoost}

Based on above evaluation and analysis, we find that ineffective cross-domain mapping is the primary factor limiting current MLLMs performance in metaphorical video understanding.
To this end, as illustrated in Figure~\ref{fig:method}, we first construct a metaphorical knowledge graph as external scaffold, then propose MetaphorBoost, a method that improves MLLMs via inference-time mapping augmentation based on the constructed metaphorical knowledge graph.

\subsection{Metaphorical Knowledge Graph}


Considering metaphor understanding typically needs interconnected  linking, we use knowledge graph for augmentation due to its intrinsic multi-hop support. 
And recognizing the need for metaphorical knowledge beyond general common sense, we construct the first metaphor-specific knowledge graph, containing 54,687 nodes and 200,268 edges.

Specifically, to construct the metaphorical knowledge graph, we first collect public textual metaphorical datasets, which contain extensive real-world metaphorical concept pairs. All texts in datasets are represented as $\mathcal{D} = \{d_1, d_2, \ldots, d_N\}$, where $N$ is  amount.
Based on this corpus, we use DeepSeek-V3.2 to extract metaphorical concept pairs from each text, which will serve as nodes in knowledge graph, as follows: 
\begin{equation}
\mathcal{C} = \bigcup_{i=1}^{N} \text{Extract}(d_i) = \bigcup_{i=1}^{N} \{(c_i^s, c_i^t)\}
\end{equation}
where $(c_i^s, c_i^t)$ are the source and target concepts with metaphorical mapping relationship, and $\mathcal{C}$ is the complete set, $|\mathcal{C}| = 54,687$.
Then we connect all obtained concepts:
\begin{equation}
\mathcal{G} = (\mathcal{C}, \mathcal{E}), \mathcal{E} = \{(c_i, c_j) \mid c_i, c_j \in \mathcal{C}, \ \text{Link}(c_i, c_j) = 1\}
\end{equation}
where $\mathcal{G}$ is the metaphorical knowledge graph, $\mathcal{E}$ is the edge set, $|\mathcal{E}| = 200,268$, $\text{Link}(\cdot, \cdot)$ indicates whether existing linking.
Detailed textual metaphorical datasets $\mathcal{D}$ are in Appendix~\ref{app:datasets}.
Prompt for extracting is in Appendix~\ref{app:extract}.

\subsection{Inference-time MetaphorVU Boosting}

Based on above metaphorical knowledge graph, we develop MetaphorBoost, aiming to consistently improve MLLMs performance via augmenting the cross-domain mapping.

\input{figs/method}

\input{tabs/ablation}

Specifically, to obtain source nodes for performing mapping augmentation, MetaphorBoost first uses given MLLM to comprehensively identify visual elements appearing in the video and output a keyword list $\mathcal{K}$, as illustrated in follows: 
\begin{equation}
\mathcal{K} = \text{Identify}(v \oplus t) = \{k_1, k_2, \ldots, k_m\}
\end{equation}
where $m$ is the amount of identified keywords in $\mathcal{K}$.
Then, MetaphorBoost queries the metaphorical knowledge graph with a maximum of $h$ hops, and retains top-$z$ target nodes that simultaneously link to the most keywords, as following:
\begin{equation}
\mathcal{R} = \text{Top-}z \left( \bigcup_{i=1}^{m} \mathcal{N}_{\mathcal{G}}^{h}(k_i), \ \text{deg}(\cdot, \mathcal{K}) \right)
\end{equation}
where $\mathcal{N}_{\mathcal{G}}^{h}(k_i)$ denotes the nodes within $h$ hops from keyword $k_i$ in metaphorical knowledge graph, $\text{deg}(\cdot, \mathcal{K})$ represents the number of edges linking a target concept to the source keywords, and $\mathcal{R}$ is the resulting set.
Finally, with retrieved concepts as reference, MetaphorBoost uses the given MLLM to reveal implicit meanings in thinking $\hat{\tau}$ and finally generate video metaphor interpretation $\hat{o}$, as follows:
\begin{equation}
\hat{\tau},\hat{o} = \text{Generate}(v \oplus t \oplus \mathcal{R})
\end{equation}
Detailed prompts for process of identifying and generating  are shown in Appendix~\ref{app:identify} and Appendix~\ref{app:generate}, respectively.

\subsection{Effectiveness of MetaphorBoost}

To extensively validate effectiveness of MetaphorBoost, we conduct experiments on multiple base models, results are in Table~\ref{tab:main_results}.
For fair comparison, MLLM settings remain consistent with baselines. For method-specific hyperparameters, number $z$ is 10, hops $h$ is 2. Main conclusion is follows. And hyperparameter experiments are in Appendix~\ref{tab:ablation_results_2}.

\textbf{MetaphorBoost can consistently improve MLLMs on metaphorical video understanding.}
As shown in Table~\ref{tab:main_results}, based on Qwen2.5-VL-7B-Instruct, average  score improve from 33.8 to 37.9 by MetaphorBoost, surpassing previous post-training methods.
Based on Qwen3-VL-8B-Thinking, average score improve from 52.0 to 55.9, surpassing previous inference-time scaling methods.
Based on Gemini-3-Pro, average score improve from 63.8 to 66.1, achieving state-of-the-art score. 
Overall, mapping augmentation via metaphorical knowledge graph can effectively and consistently boosts MLLMs on metaphorical video understanding.

\subsection{Ablation of MetaphorBoost}

To further explore, we conduct ablation on introducing external knowledge, constructing graph structure, and using metaphor-oriented knowledge in Table~\ref{tab:ablation_results}. Conclusions are: 

\textbf{External knowledge is important for mapping augmentation.} 
``w/o external augmentatio'' means querying the MLLM itself for augmentation instead of using external knowledge. The performance drops compared to MetaphorBoost, indicating that external knowledge helps compensate for MLLMs deficiency in the cross-domain mapping.

\textbf{Knowledge graph provides more effective augmentation than plain text.} 
``w/o graph-structure augmentation'' means retrieving from raw textual metaphorical datasets instead of querying the knowledge graph. The performance drop demonstrates that graph structures provide more effective mapping augmentation by explicit relational connections.

\textbf{Metaphor-oriented augmentation outperforms commonsense augmentation.} 
``w/o metaphor-oriented augmentation'' means using ConceptNet\footnote{{https://huggingface.co/spaces/cstr/conceptnet\_db}}, a general commonsense knowledge graph, instead of our metaphorical knowledge graph. Performance drops, further supporting  MetaphorVU requires the high-order cognition beyond basic knowledge.

\input{figs/error}

\subsection{Detailed Analysis for MetaphorBoost}

\textbf{Decline of Bad Mapping Amount.}
To further reveal why MetaphorBoost achieves the performance improvement, we analyze thinking process of MetaphorBoost and count the occurrences of missing, superficial and improper mapping, and compare with base models. As shown in Figure~\ref{fig:error}, the reduced amount of missing, superficial and improper mapping confirms that \textit{MetaphorBoost effectively boosts metaphorical video understanding by enhancing the capability linking visual elements to external underlying concepts}.

\textbf{Case Study.}
To provide more concrete illustration of reasons why MLLMs struggle with metaphorical video understanding, as well as how MetaphorBoost improves performance, we present a representative case study. As shown in Appendix Figure~\ref{fig:case_study}, 
the green, orange, and blue highlights indicate missing mapping, superficial mapping, and improper mapping respectively, collectively leading to poor metaphorical video interpretation. And \textit{MetaphorBoost effectively mitigates the three types of deficiencies, thereby improving MLLMs performance on metaphorical video understanding}.

%% file: figs/method.tex
\begin{figure}[t!]
\centering
\includegraphics[width=\linewidth]{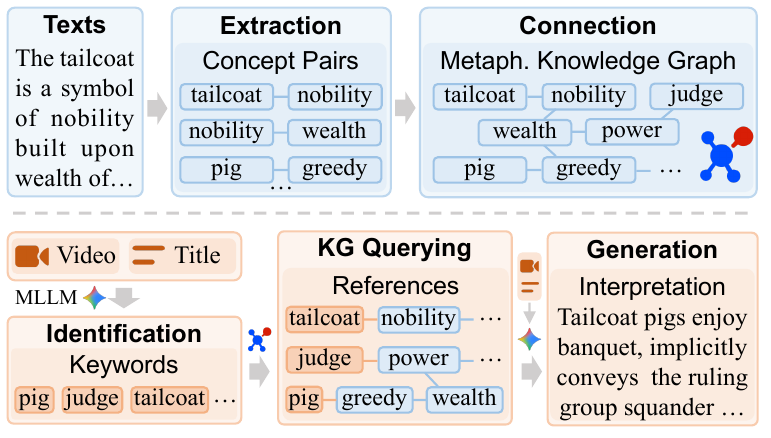} 
\caption{We construct a metaphorical knowledge graph and then propose MetaphorBoost, improving MLLMs performance on metaphorical video understanding via mapping augmentation.}
\label{fig:method}
\end{figure}

%% file: tabs/ablation.tex
\begin{table*}[t!]

\caption{Ablation results show that external knowledge is important for mapping augmentation,  structured knowledge graph provides more effective augmentation than plain text, and augmentation by metaphor-oriented knowledge outperforms commonsense knowledge.}

\centering
\resizebox{\linewidth}{!}{
\begin{tabular}{lccccccccc}
\toprule

\textbf{Method} & \textbf{Body L.} & \textbf{Atmosph. L.} & \textbf{Cultural S.} & \textbf{Natural. S.} & \textbf{Causal M.} & \textbf{Analog. M.} & \textbf{Surreal N.} & \textbf{Perform. N.} & \textbf{Average} \\
  

\midrule

MetaphorBoost (Qwen3-VL-8B-Thinking) (Ours)~~~~~ & \textbf{61.8} & \textbf{71.0} & \textbf{71.8} & \textbf{61.3} & \textbf{36.7} & \textbf{47.1} & \textbf{45.7} & 31.5 & \textbf{55.9} \\

~~w/o external augmentation & 57.1 & 69.9 & 67.6 & 60.3 & 33.9 & 44.9 & 40.5 & \textbf{36.6} & 53.4 \\

~~w/o graph-structure augmentation & 60.5 & 70.3 & 69.8 & 61.0 & 30.0 & 43.3 & 45.5 & 30.8 & 54.3 \\

~~w/o metaphor-oriented augmentation & 57.3 & 67.5 & 65.6 & 61.0 & 30.0 & 46.0 & 42.2 & 30.0 & 52.5 \\

\bottomrule
\end{tabular}%
}
\label{tab:ablation_results}
\end{table*}

%% file: figs/error.tex
\begin{figure}[t!]
\centering
\includegraphics[width=\linewidth]{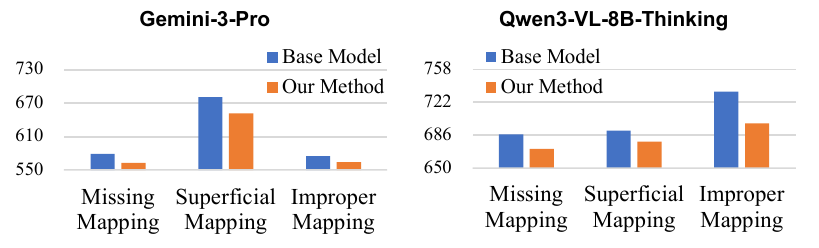} 
\caption{Amount of three kinds of bad mapping reduces, proving MetaphorBoost can effectively enhance cross-domain mapping.}
\label{fig:error}
\end{figure}

%% file: sections/related.tex
\section{Related Work}

\paragraph{Metaphor Understanding.}
Prior research on metaphor understanding primarily focuses on text and images, with video metaphor remaining relatively scarce.
For textual metaphor, works aim to detect metaphor based on relationships between tokens, and to identify the source and target domains~\cite{prystawski2023psychologically,tian2024bridging,zheng2025enhancing}. 
For image metaphor, some works collect images such as internet memes for datasets~\cite{xu2022met,yang2025cultural,kundu2025looking,saakyan2025understanding,chakrabarty2022flute}, or explore multimodal fusion to improve performance~\cite{qian2025concept,zheng2025multi,xu2024exploring}.
Compared to text and images, videos are temporal and convey richer information, more likely containing complex metaphor. 
Recently, a few studies advance video metaphor research by constructing datasets from advertisement videos~\cite{kalarani2024unveiling,jia2025summa,long2025adsqa,zhang2025videoads}. However, these are limited to the advertising domain, which may not accurately reflect the capabilities in complex real-life scenarios.

\textbf{Deep-semantic Video Understanding.}
With the advancement of MLLMs, recent work begins to explore deep-level video understanding beyond basic object recognition or event description.
Some studies present scientific experiment in videos and require to predict outcomes~\cite{deng2025scivideobench},
illustrate complex domain knowledge and require  to solve new problems not shown in the video~\cite{hu2025video},
show incomplete event and ask to infer the underlying logic of event~\cite{chen2025looking},
and display objects from the same scene across separate frames, requiring to reason about spatial relationships and motion trajectories~\cite{swetha2025implicitqa,yang2025thinking}.
Additionally, some studies investigate advertisement video understanding, as discussed in above paragraph.
Overall, research on deep-semantic video understanding remains in the early stages. Our work contributes to this direction by systematically introducing metaphorical video understanding as a new challenging task.

Recently, \textbf{MMR-V} is proposed to evaluate the implicit reasoning in video understanding~\cite{zhu2026mmrv}, which is a highly valuable related work.
Upon careful comparison, the core unique value of our work lies in the systematicness and depth on metaphorical video understanding compared with MMR-V.
Specifically, MMR-V aims to assess a broad spectrum of reasoning abilities, where metaphor-related content appears as one of many test scenarios rather than a dedicated focus. In contrast, our work focuses specifically on metaphorical video understanding. We construct a systematic taxonomy of video metaphor and carefully curate a benchmark spanning diverse metaphor types and topics, thereby enabling more comprehensive and fine-grained analysis of MLLMs' metaphorical video understanding capability.
From a broader perspective, our work and MMR-V can complement each other, jointly enabling a deep evaluation of high-order cognitive capabilities to improve MLLMs. 

\textbf{Multimodal Sarcasm.} Multimodal sarcasm research is relevant to our work and deserves discussion. In general, multimodal sarcasm understanding and metaphorical video understanding differ in their core capability requirements and the types of implicit meanings they encompass.
In terms of core capability requirements, sarcasm primarily relies on identifying apparent contradictions among elements~\cite{zhuang2025multi,wang2025can}, whereas metaphorical video understanding requires models to perform cross-domain mapping, i.e., linking visual elements to underlying concepts.
In terms of implicit meanings, sarcasm mainly focuses on conveying critical and negative thoughts~\cite{wang2025s3,ou2025multi}, whereas metaphorical video understanding covers a broader and more diverse range of implicit meanings, as in Figure~\ref{fig:bench}, encompassing various forms prevalent in everyday life.

%% file: sections/conclusion.tex
\section{Conclusion}

In this paper, to fill the gap in prior research on metaphorical video understanding, 
we design the first systematic video metaphor taxonomy and construct MetaphorVU-Bench, enabling a comprehensive evaluation of metaphorical video understanding.
Extensive experiments reveal that current MLLMs struggle with accurate metaphorical video understanding, primarily due to defective cross-domain mapping.
Motivated by these findings, we construct a metaphorical knowledge graph and propose MetaphorBoost, which can consistently improve MLLM performance via mapping augmentation.  
This paper offers a promising direction for MLLM advancement and can inspire further research.

%% file: apps/theory.tex
\section{Theoretical Basis for Video Metaphor Taxonomy}
\label{app:theory}

To ensure reliable and principled evaluation, a systematic video metaphor taxonomy is essential for building the benchmark. 
Since no prior works have explored this kind of taxonomy, we draw on multimodal metaphor theory~\cite{forceville2009non,forceville2009multimodal} and its extensions in the video field~\cite{bordwell2013viewer,stam2017film,schechner2017performance,chandler2022semiotics}, designing the first systematic video metaphor taxonomy, the details are illustrated in follows:

According to Film Mise-en-scène Theory~\cite{bordwell2004film, gibbs2002mise, arnheim1957film}, video metaphors can be realized through visual element arrangement within frames. \textbf{Body Language} corresponds to Performance Staging—physical movements, facial expressions, and postures serve as metaphorical source domains, mapping abstract emotional states onto visible bodily behaviors~\cite{naremore1988acting, gibbs2002mise}. \textbf{Atmosphere Language} corresponds to Environmental Staging—color tones, lighting, and composition serve as metaphorical carriers of emotional tone~\cite{arnheim1957film, bellantoni2012if, brown2016cinematography}.

According to Symbol and Symbolism Theory~\cite{rawls1999collected, jung2012man, eliade1991images, chandler2022semiotics}, video metaphors can be realized through symbolic signs carrying conventional or archetypal meaning. \textbf{Cultural Symbol} corresponds to conventionally established symbols within specific cultural contexts—their meaning depends on cultural knowledge~\cite{danesi2018cigarettes, yeats1998mythologies}. \textbf{Naturalistic Symbol} corresponds to natural elements with universal symbolic meaning rooted in shared human experiences and collective unconscious~\cite{jung2012man, campbell2008hero, ferber1999dictionary}.

According to Montage Theory~\cite{eisenstein2018film, kuleshov1974kuleshov, pudovkin2013film, cutting2016narrative}, video metaphors can be realized through dialectical collision between shots. \textbf{Causal Montage} corresponds to causal reasoning—temporal shot juxtaposition implies causal relationships, with audiences automatically completing causal chains~\cite{pudovkin2013film, bordwell2013narration, carroll1996theorizing}. \textbf{Analogical Montage} corresponds to analogical reasoning—juxtaposition of unrelated shots guides audiences to identify structural similarities and establish cross-domain mappings~\cite{eisenstein2018film, whittock1990metaphor, fauconnier2008way}.

According to Theatre Semiotics and Performance Theory~\cite{elam2003semiotics, schechner2017performance}, narrative-based video metaphors operate through distinct semiotic registers. \textbf{Surreal Narrative} employs what  terms ``virtual performance''—animated or AI-generated characters transcend physical constraints, enabling metaphorical expression through impossible actions, fantastical transformations, and dreamlike scenarios that would be unachievable in reality~\cite{auslander2022liveness, manovich2002language, wells2013understanding}. \textbf{Performative Narrative} relies on embodied performance where human actors serve as direct meaning carriers; audiences decode metaphorical connotations through theatrical conventions such as exaggerated expressions, symbolic staging, and dramatized conflicts~\cite{schechner2017performance, elam2003semiotics}.

%% file: apps/construction.tex
\section{Multi-stage Filtration Prompts and Manual Annotation Guideline}

\subsection{Prompt for LLM Filtration}
\label{app:filter_llm}
To efficiently isolate metaphorical videos from billions of videos, we first use a powerful LLM (GPT-5) to analyze the video introduction, automatic speech recognition (ASR) result and audience comments to determine whether each video contains metaphorical logic, the detailed prompt is shown in Figure~\ref{fig:prompt_for_LLM_fil}.

\subsection{Prompt for MLLM Filtration}
\label{app:filter_mllm}
Considering above filtration process does not directly use
visual information and LLM analysis may not align with
the actual video, to conduct further check and filtration, a powerful MLLM (Gemini-3-Pro) is used to verify whether above analysis is consistent with original videos, the detailed prompt is shown in Figure~\ref{fig:prompt_for_MLLM_fil}.

\subsection{Prompt for Human Filtration}
\label{app:filter_human}
Then, a human team performs final filtration based on the original video, video introduction and audience comments, resulting in 860 videos with definite metaphorical logic. 
Additionally, annotators identify the metaphor type for each video, balancing the number of samples across each metaphor type as much as possible. The detailed prompt is shown in Figure~\ref{fig:prompt_for_Human_fil}.

\subsection{Manual Annotation Guideline}
\label{app:annotation}
When annotating video metaphor interpretation, we require human annotators to reference video introduction and audience comments and follow a fixed format (i.e., \textit{specifying which visual elements convey which implicit meanings}). The detailed guideline is shown in Figure~\ref{fig:prompt_for_Human_ann}.

%% file: apps/evaluation.tex
\section{Prompt for Evaluation and LLM Judge, and Consistency Experiments}

\subsection{Prompt for Evaluation}
\label{app:evaluation}
Generally, MLLMs first recognize visual contents, establish projection to external concepts and unveil implicit meanings in thinking process, then interpret which visual contents
convey which implicit meanings in final output. Details of evaluation prompt  are in Figure~\ref{fig:prompt_for_evalu}.

\subsection{Prompt for LLM Judge}
\label{app:judge}

Since the output  video metaphor interpretation in MetaphorVU-Bench is free-form text, rule-based metrics are difficult to provide a score aligning with actual human habits~\cite{mayfield2024evaluation,li2025benchmark}. To this end, we follow the metrics in previous free-form QA evaluation works~\cite{li2024structrag,li2025deepsolution,yu2025vrbench,long2025adsqa}, using DeepSeek-V3.2 as LLM judge. Detailed prompt for LLM judge are in Figure~\ref{fig:prompt_for_judge}.

\subsection{Consistency Experiments for LLM Judge}
\label{app:consistency}

To verify the reliability of the LLM judge, we randomly sample 100 instances from the evaluation results and have human annotators score the model-generated video metaphor interpretations following the same evaluation guidelines. We then analyze the consistency between human scores and LLM judge scores. The results show a Pearson correlation coefficient of 0.85 with a p-value of 3e-20 ($p < 0.001$), indicating a strong positive correlation with high statistical significance between human and LLM judgments. This validates the reliability and effectiveness of using LLM as an automatic judge in our framework.

%% file: apps/extract_and_datasets.tex
\section{Textual Datasets and Prompt in Metaphorical KG Construction}

\subsection{Details of Metaphorical Textual Datasets}
\label{app:datasets}
To construct a metaphorical knowledge graph, we first collect textual metaphorical datasets, which contain extensive metaphorical concept pairs. The details of used textual metaphorical datasets are shown in Table~\ref{tab:datasets_tab}.
Note that a portion of the data was originally in Chinese, to ensure the universality of the metaphorical knowledge graph, we use GPT-5 to translate the original text into English.
\begin{table}[t]
\caption{Details of metaphorical textual datasets.}
\centering
\resizebox{\linewidth}{!}{
\begin{tabular}{lcc}
\toprule
{\textbf{Name}} & \textbf{URL} & \textbf{\# Samples} \\
\midrule
Manual\_Metaphors & \url{https://huggingface.co/datasets/Sasidhar1826/manual_data_on_metaphors} & 718   \\
Metaphor\_Novelty & \url{https://huggingface.co/datasets/omarmomen/metaphor-novelty}  & 200 \\
Metaphor\_Explanation & \url{https://huggingface.co/datasets/JasonShao/Chinese_Metaphor_Explanation}  & 28000 \\
Metaphor\_Dataset & \url{https://huggingface.co/datasets/liyucheng/chinese_metaphor_dataset}  & 8030 \\
\bottomrule
\end{tabular}%
}
\label{tab:datasets_tab}
\end{table}%

\subsection{Prompt for Extracting Metaphorical Concept Pairs}
\label{app:extract}
Since several previous works that have been widely recognized by the community have demonstrated that current LLMs possess excellent information extraction capabilities~\cite{tang2023harvesting,li2024meta,tang2024self,li2025paperregister}, we adopt the same approach and use DeepSeek-V3.2 to extract metaphorical concept pairs from each text, which will serve as nodes in the knowledge graph. The specific prompt is shown in Figure~\ref{fig:prompt_for_extract_pair}.

%% file: apps/identify_and_generate.tex
\section{Prompts for Identification and Generation in MetaphorBoost}

\subsection{Prompt for Identifying Visual Elements}
\label{app:identify}
At the time of MLLMs inference, to obtain the source nodes for performing cross-domain mapping augmentation, MetaphorBoost first uses the given MLLM to comprehensively identify visual elements appearing in the video and output a keyword list. The specific prompt is shown in Figure~\ref{fig:prompt_for_identi}.

\subsection{Prompt for Generating Video Metaphor Interpretation}
\label{app:generate}
Based on above identifying results, MetaphorBoost queries the metaphorical knowledge graph. And then with retrieved concepts as augmentation, MetaphorBoost uses the given MLLM to unveil implicit meanings and finally generate video metaphor interpretation. The specific prompt is shown in Figure~\ref{fig:prompt_for_gene}.

%% file: apps/baselines.tex
\section{Details of Reasoning-based Baselines}
\label{app:baselines}

Reasoning-enhanced Methods improve the reasoning ability of base model by post-training or inference-time scaling, this type of baseline includes 7 methods:


\textbf{VideoRFT}~\cite{wang2025videorft} is a reinforcement fine-tuning approach designed to cultivate video reasoning capabilities in multimodal large language models. It follows a two-stage training scheme: supervised fine-tuning with chain-of-thought annotations, followed by reinforcement learning with a semantic-consistency reward to promote alignment between textual reasoning and visual evidence. While VideoRFT achieves strong performance on various video reasoning benchmarks, it primarily focuses on foundational cognitive tasks such as object recognition and event understanding, limiting its capability for metaphorical video understanding.

\textbf{Vision-R1}~\cite{huang2025vision} aims to enhance multimodal reasoning capability through reinforcement learning inspired by DeepSeek-R1. It constructs a 200K multimodal CoT dataset via modality bridging and data filtering, and employs Progressive Thinking Suppression Training to refine complex reasoning ability. However, similar to VideoRFT, it is primarily tailored for low-level video understanding tasks involving logical and mathematical reasoning, rather than the cross-domain mapping required for metaphorical video interpretation.

\textbf{ReAd-R}~\cite{long2025adsqa} is a reinforcement learning model specifically designed for advertisement video understanding, targeting tasks that require perceiving beyond objective physical content, such as marketing logic and persuasive strategies. Compared to VideoRFT and Vision-R1, ReAd-R is more relevant to our task as advertisement videos often contain implicit meanings. However, its domain-specific training limits generalizability to broader metaphorical video understanding.

\textbf{LTR}~\cite{liaodivide} (Language-centric Tree Reasoning) enhances video question-answering through structured logical reasoning at inference time. It recursively divides complex cognitive questions into manageable parts and performs bottom-up reasoning within a language-centric logical tree. While LTR improves reasoning transparency on various video QA benchmarks, its structured decomposition approach may not effectively capture the cross-domain mapping required for understanding video metaphors.

\textbf{ViTCoT}~\cite{zhang2025vitcot} (Video-Text Interleaved Chain-of-Thought) introduces a video reasoning paradigm that interleaves visual and textual information during reasoning, enabling models to re-examine visual content while reasoning. Although ViTCoT improves general video understanding by better integrating visual modality, it still focuses on explicit content reasoning rather than cross-domain mapping required in metaphorical understanding.

\textbf{Prompt Engineering}~\cite{wei2022chain} refers to chain-of-thought prompting, which improves reasoning ability by generating intermediate reasoning steps through carefully designed prompts. In our experiments, we design prompts that explicitly encourage the model to perform cross-domain mapping from visual contents to implicit meanings, representing a straightforward baseline for metaphorical video understanding.

\textbf{Few-shot Example}~\cite{dong2024survey} is based on in-context learning, where models make predictions based on contexts augmented with demonstration examples. For metaphorical video understanding, we provide annotated examples demonstrating how to project explicit visual contents onto abstract concepts. Together with Prompt Engineering, this represents the most direct approach for adapting existing models to our task.

%% file: apps/hyperpara.tex
\section{Experiments about Query Strategy and Hyperparameters}
\label{app:hyperpara}

In the inference-time mapping augmentation, MetaphorBoost queries the metaphorical knowledge graph with a maximum of $h$=2 hops, and retains the Top-$z$=10 target nodes that are simultaneously associated to the most keywords, thereby maximizing the advantages of the knowledge graph, namely its support for multi-hop and structured reasoning. 
To convincingly demonstrate the effectiveness of this query strategy, we conduct further experiments, as shown in Table~\ref{tab:ablation_results_2}. 

The setting ``w/o common connection'' means that instead of retaining results that simultaneously have as many connections to the query keywords as possible, results are retained randomly. 
The experimental results show that the average performance decreases. This, to some extent, \textit{demonstrates the advantages of using a knowledge graph, which can provide low-noise augmentation via structured federated query}.

Furthermore, to provide a deeper investigation into the underlying mechanism of MetaphorBoost, we conduct experiments on its two key hyperparameters: the maximum number of hops $h$ for querying the knowledge graph and the number of retained results $z$, with default values of 2 and 10, respectively. In the table, we present results for $h=1$ and $z=5$. The experimental results show that while performance fluctuates across different subsets, the average scores of all variants remain lower than those of MetaphorBoost with default settings. This further validates the effectiveness of leveraging the knowledge graph for cross-domain mapping—demonstrating that \textit{the knowledge graph can provide effective, reasonably deep, and low-noise augmentation for metaphor interpretation}.

\begin{table*}[t!]
\caption{Experiments about query strategy and hyperparameters.}

\centering
\resizebox{\linewidth}{!}{
\begin{tabular}{lccccccccc}
\toprule

\textbf{Method} & \textbf{Body L.} & \textbf{Atmosph. L.} & \textbf{Cultural S.} & \textbf{Natural. S.} & \textbf{Causal M.} & \textbf{Analog. M.} & \textbf{Surreal N.} & \textbf{Perform. N.} & \textbf{Average} \\
  
\midrule

MwtaphorBoost (Qwen3-VL-8B-Thinking) & \textbf{61.8} & 71.0 & 71.8 & 61.3 & \textbf{36.7} & 47.1 & \textbf{45.7} & 31.5 & \textbf{55.9} \\
\midrule
~~w/o common connection & 59.5 & 69.7 & \textbf{72.3} & 62.2 & 35.0  & 45.3 & 43.7 & 33.5 & 54.8 \\
\midrule
~~w/ hop $h=1$, return $z=10$ & 59.3 & \textbf{73.0}  & 68.5  & \textbf{65.4} & 25.3 & 46.4 & 42.9 & 32.5 & 54.5 \\
~~w/ hop $h=2$, return $z=5$ & 60.1 & 71.8 & 70.0 & 63.7 & 31.3 & \textbf{47.5} & 45.1 & \textbf{35.6} & 55.7 \\

\bottomrule
\end{tabular}%
}
\label{tab:ablation_results_2}
\end{table*}

%% file: apps/case.tex
\section{More Examples of MetaphorVU-Bench}
\label{app:case}

We provide more examples for all eight video metaphor types, specifically, Body Language is in Figure~\ref{fig:more_case_1_fig}, Atmosphere Language is in Figure~\ref{fig:more_case_2_fig}, Cultural Symbol is in Figure~\ref{fig:more_case_3_fig}, Naturalistic Symbol is in Figure~\ref{fig:more_case_4_fig}, Causal Montage is in Figure~\ref{fig:more_case_5_fig}, Analogical Montage is in Figure~\ref{fig:more_case_6_fig}, Surreal Narrative is in Figure~\ref{fig:more_case_7_fig}, Performative Narrative is in Figure~\ref{fig:more_case_8_fig}.

%% file: figs/case_study.tex
\begin{figure*}[t!]
\centering
\includegraphics[width=\linewidth]{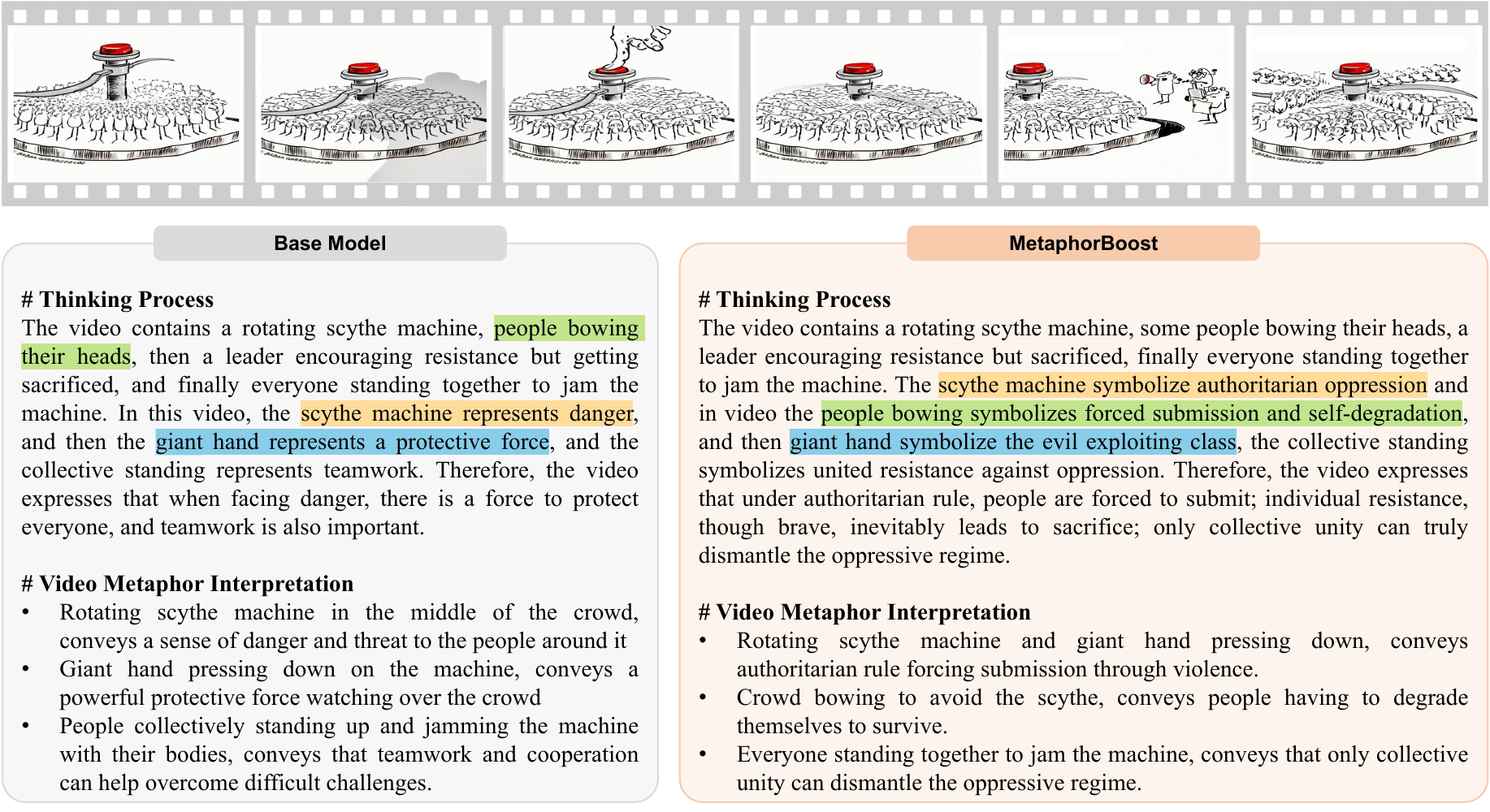} 
\caption{The green, orange, and blue highlights indicate missing mapping, superficial mapping, and improper mapping respectively, these deficiencies collectively lead to poor metaphorical video interpretation. MetaphorBoost effectively mitigates the three types of deficiencies, thereby improving MLLMs performance on metaphorical video understanding.}  
\label{fig:case_study}
\end{figure*}

%% file: figs/prompt_for_LLM_fil.tex
\begin{figure}[t!]
\centering
\includegraphics[width=\linewidth]{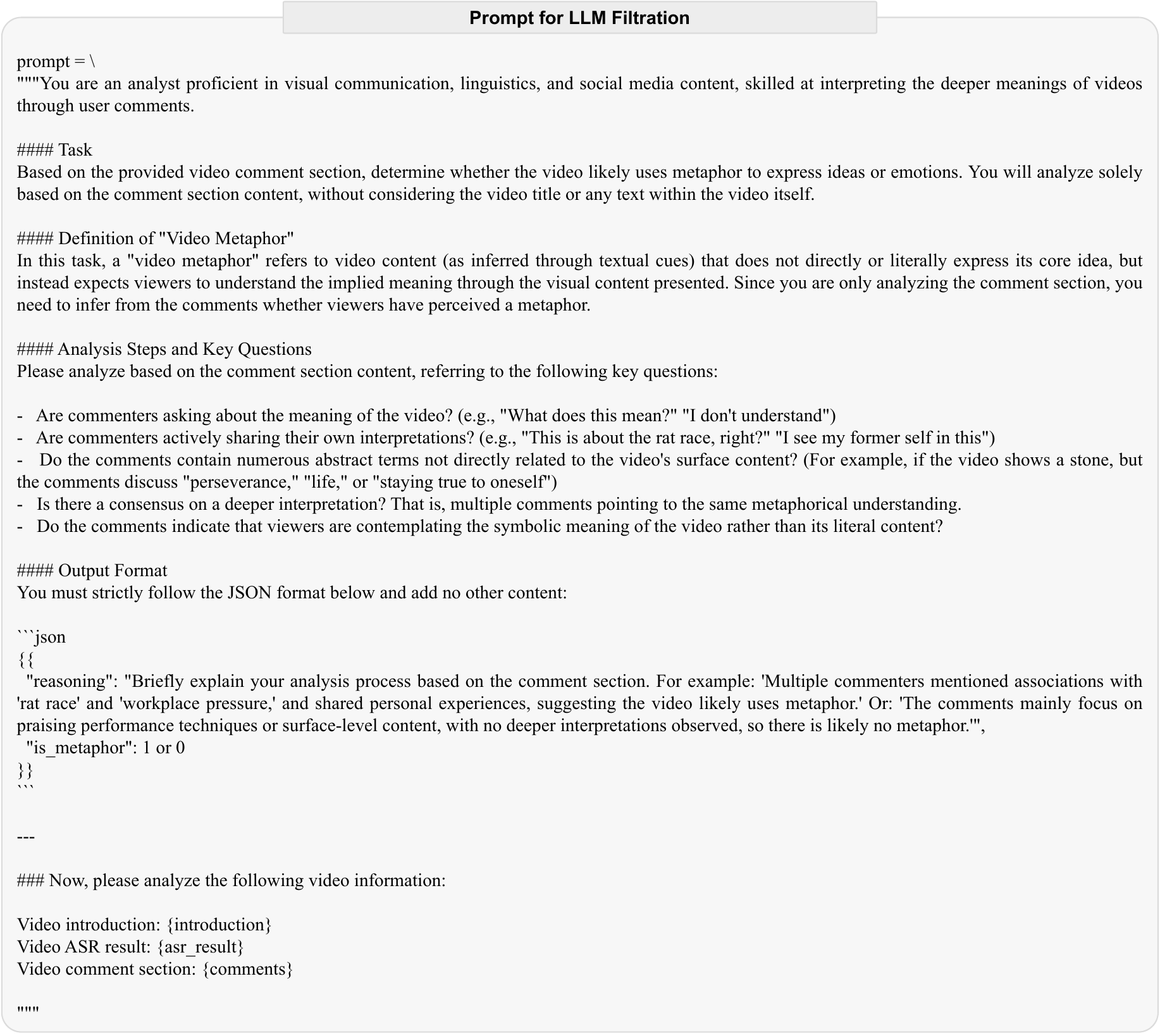} 
\caption{Prompt for LLM filtration.}
\label{fig:prompt_for_LLM_fil}
\end{figure}

%% file: figs/prompt_for_MLLM_fil.tex
\begin{figure}[t!]
\centering
\includegraphics[width=\linewidth]{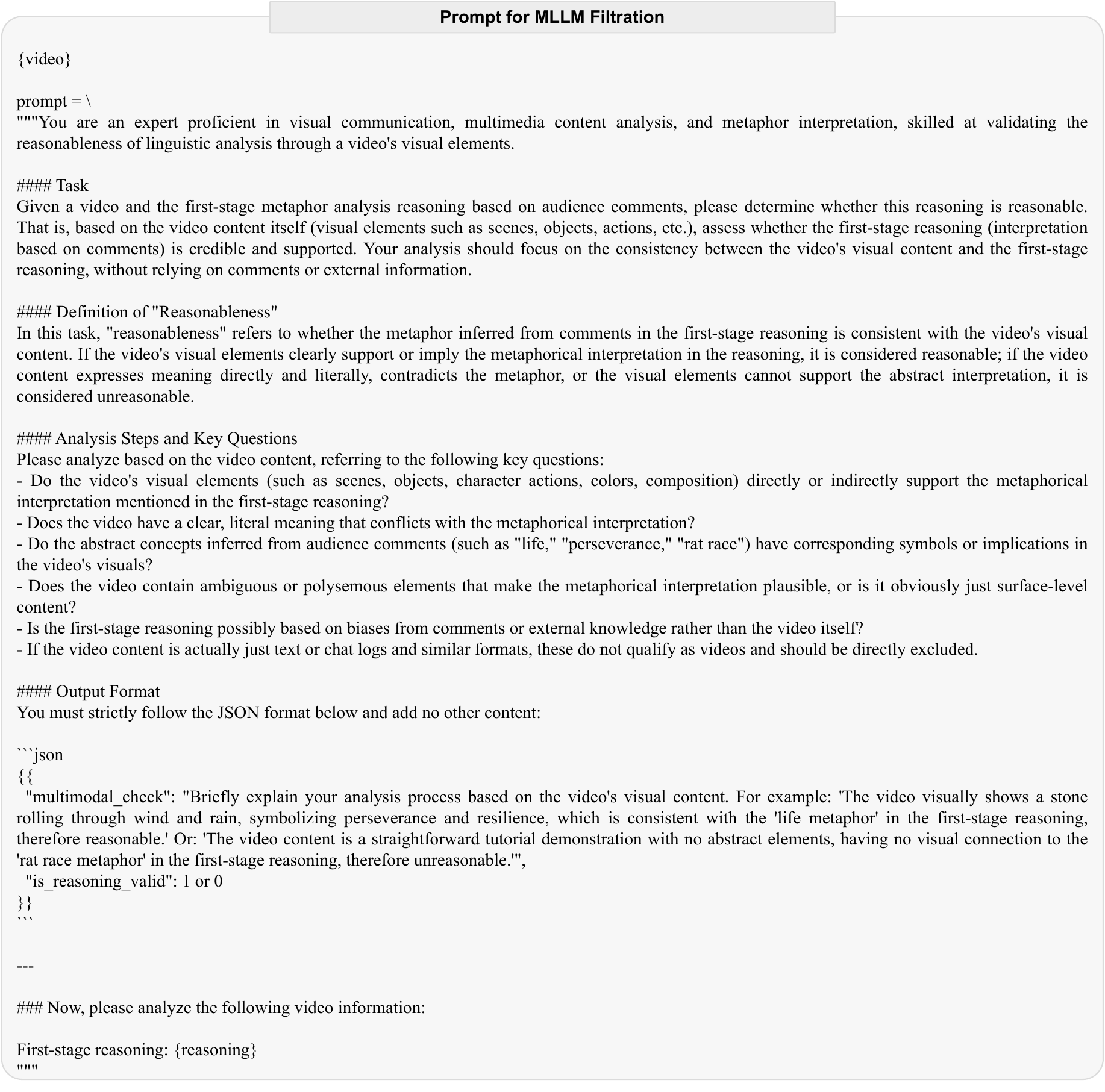} 
\caption{Prompt for MLLM filtration.}
\label{fig:prompt_for_MLLM_fil}
\end{figure}

%% file: figs/prompt_for_human_fil.tex
\begin{figure}[t!]
\centering
\includegraphics[width=\linewidth]{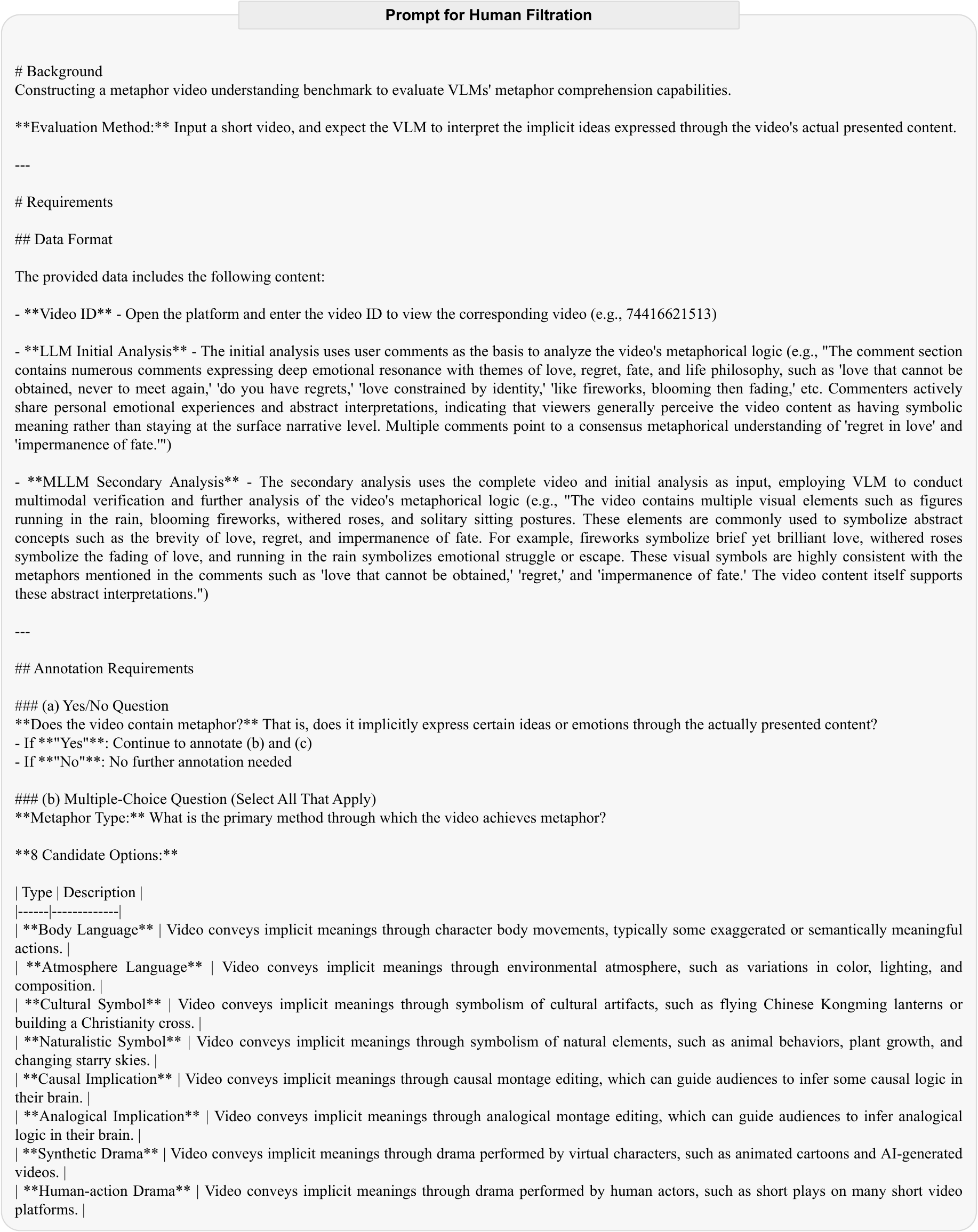} 
\caption{Prompt for Human filtration.}
\label{fig:prompt_for_Human_fil}
\end{figure}

%% file: figs/prompt_for_human_ann.tex
\begin{figure}[t!]
\centering
\includegraphics[width=\linewidth]{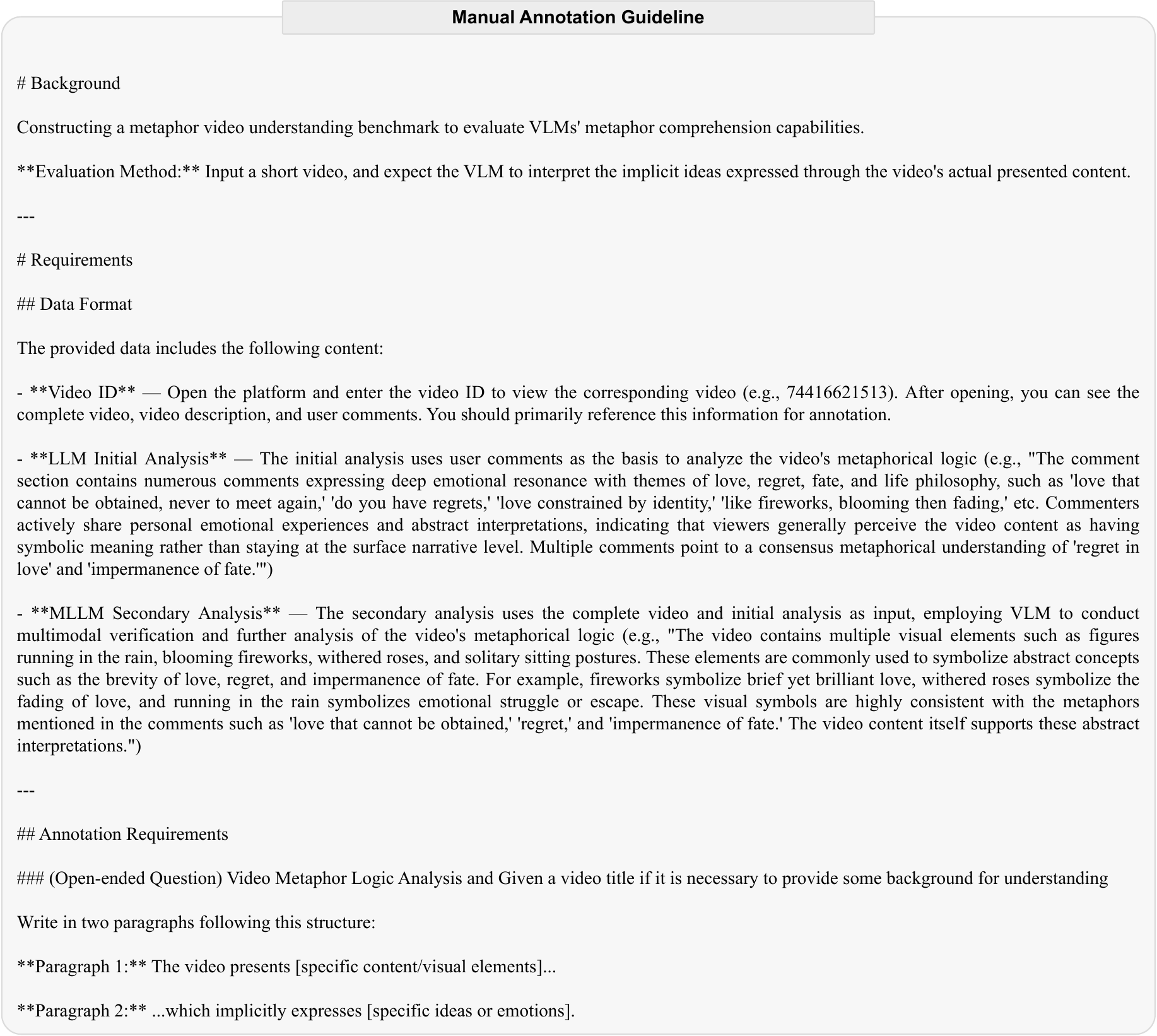} 
\caption{Manual annotation guideline.}
\label{fig:prompt_for_Human_ann}
\end{figure}

%% file: figs/prompt_for_evalu.tex
\begin{figure}[t!]
\centering
\includegraphics[width=\linewidth]{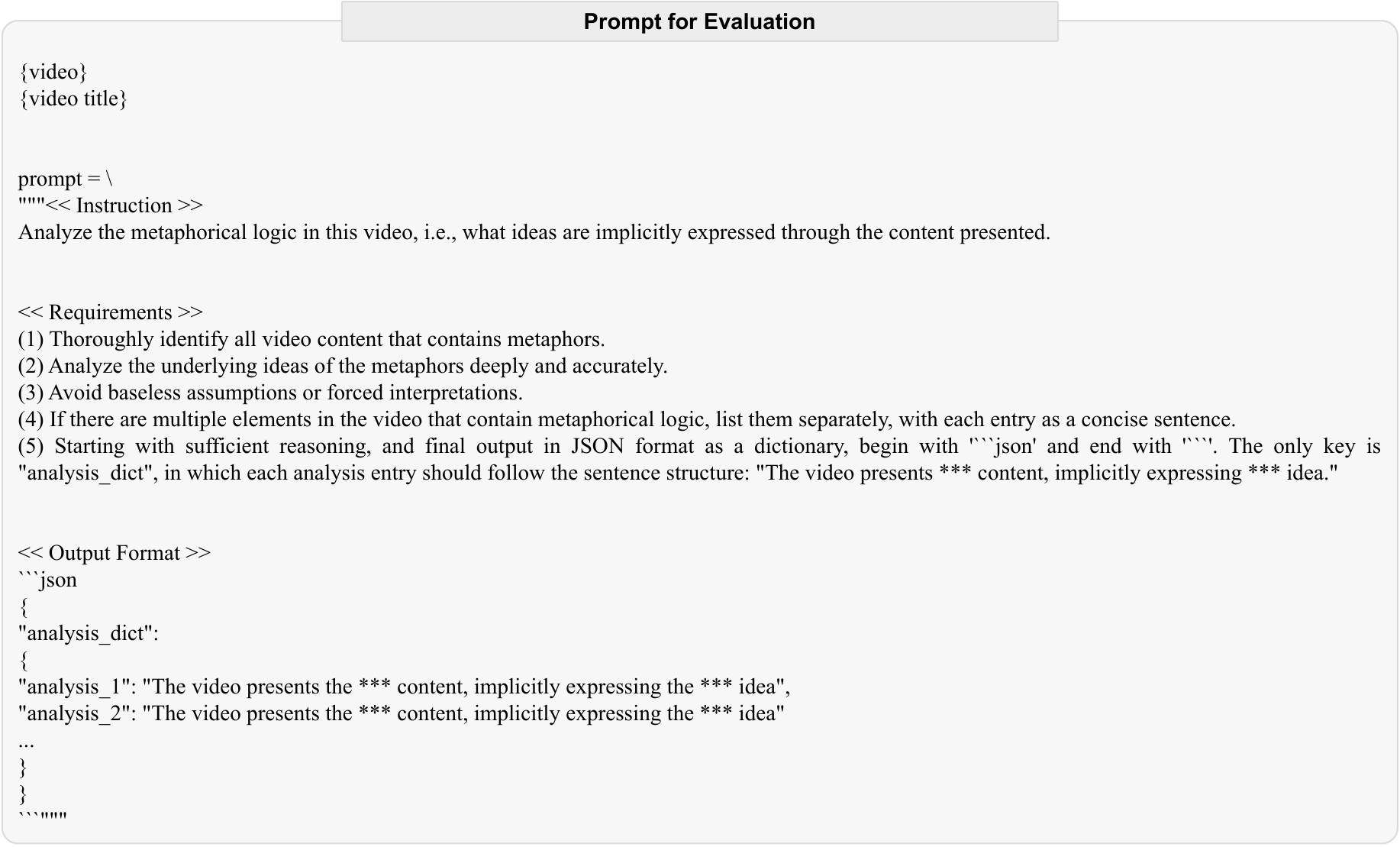} 
\caption{Prompt for evaluation.}
\label{fig:prompt_for_evalu}
\end{figure}

%% file: figs/prompt_for_judge.tex
\begin{figure}[t!]
\centering
\includegraphics[width=\linewidth]{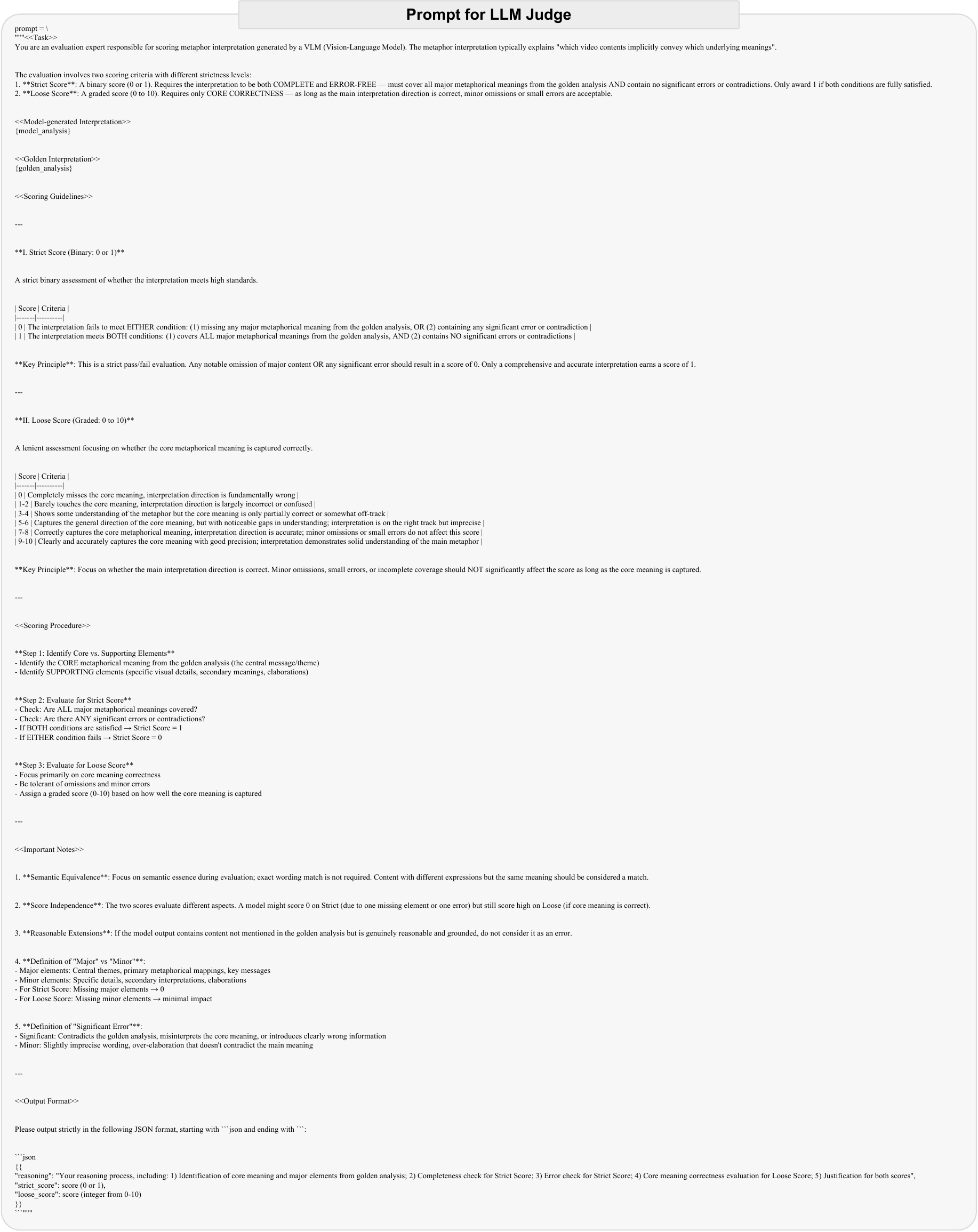} 
\caption{Prompt for LLM judge.}
\label{fig:prompt_for_judge}
\end{figure}

%% file: figs/prompt_for_extract_pair.tex
\begin{figure}[t!]
\centering
\includegraphics[width=\linewidth]{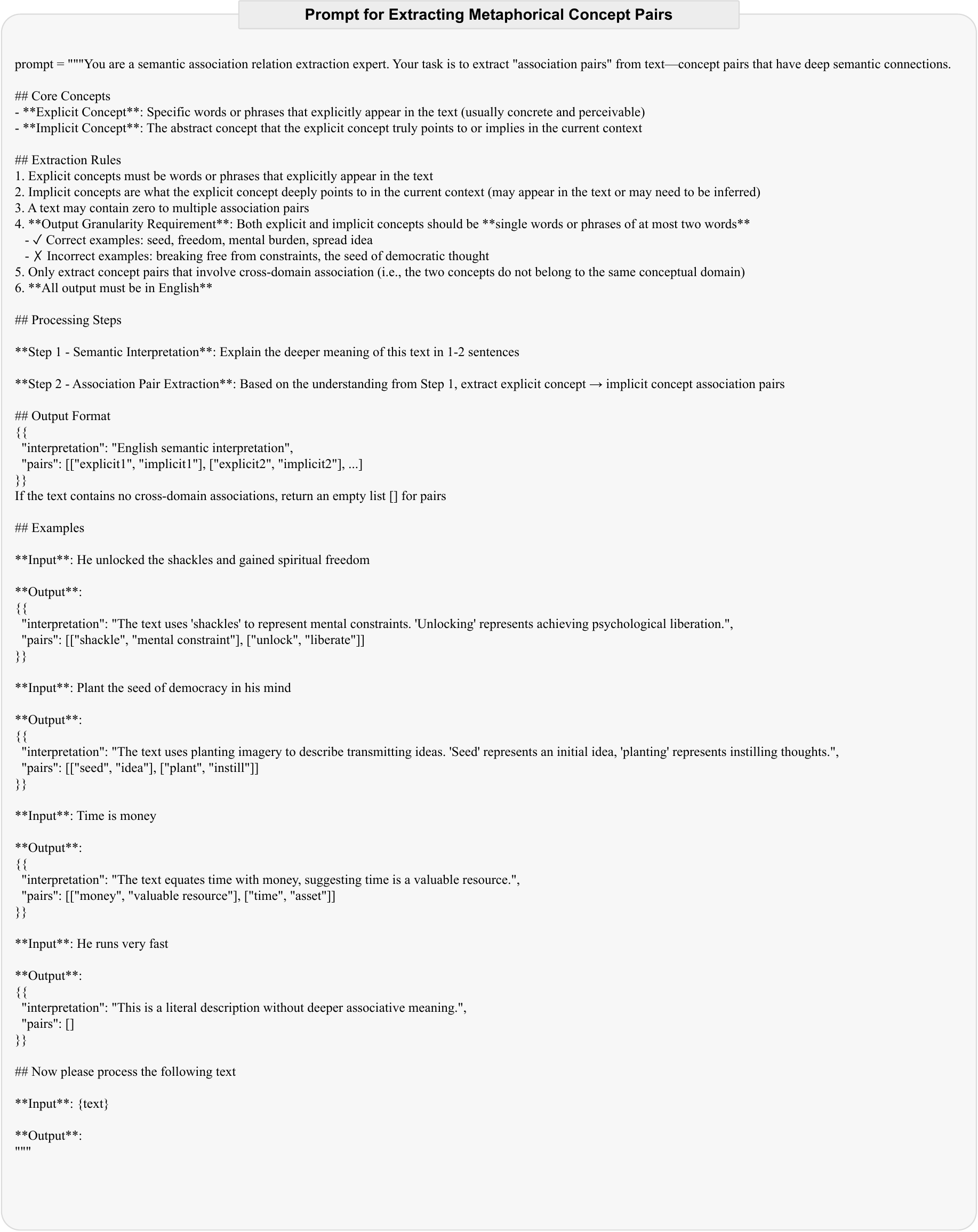} 
\caption{Prompt for extracting metaphorical concept pairs.}
\label{fig:prompt_for_extract_pair}
\end{figure}

%% file: figs/prompt_for_identi.tex
\begin{figure}[t!]
\centering
\includegraphics[width=\linewidth]{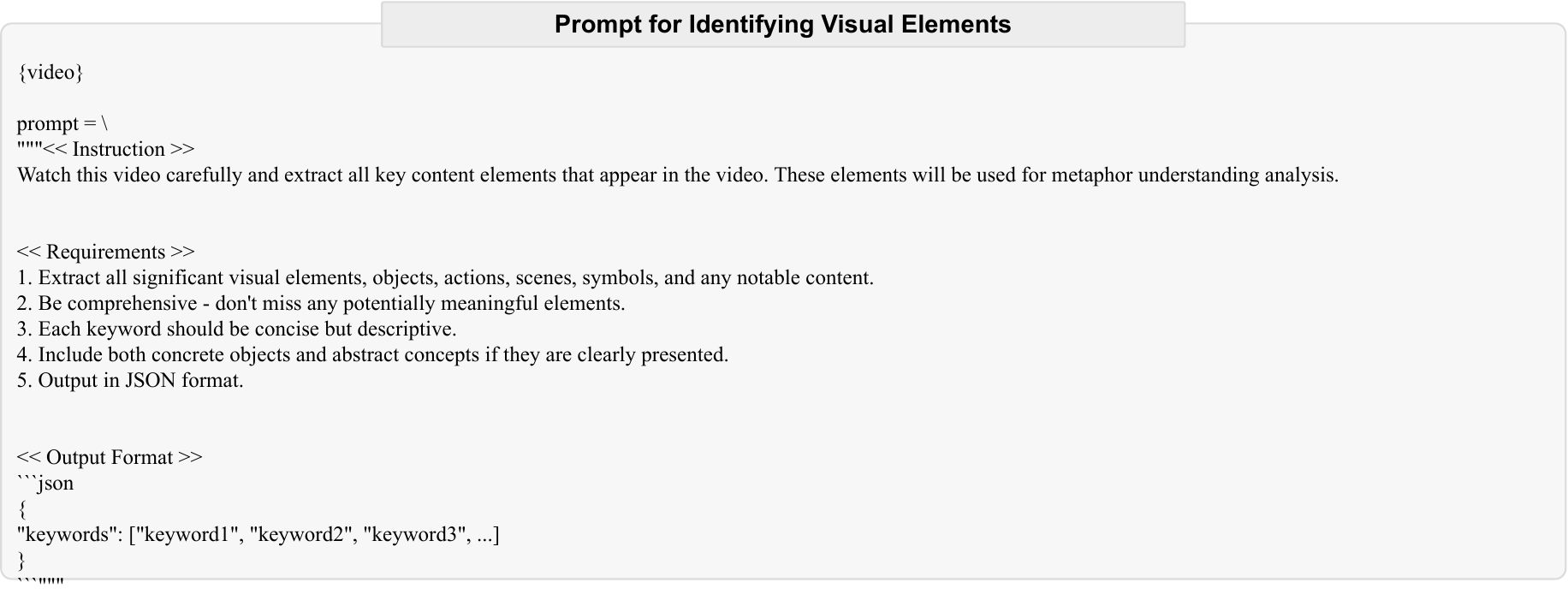} 
\caption{Prompt for identifying visual elements.}
\label{fig:prompt_for_identi}
\end{figure}

%% file: figs/prompt_for_gene.tex
\begin{figure}[t!]
\centering
\includegraphics[width=\linewidth]{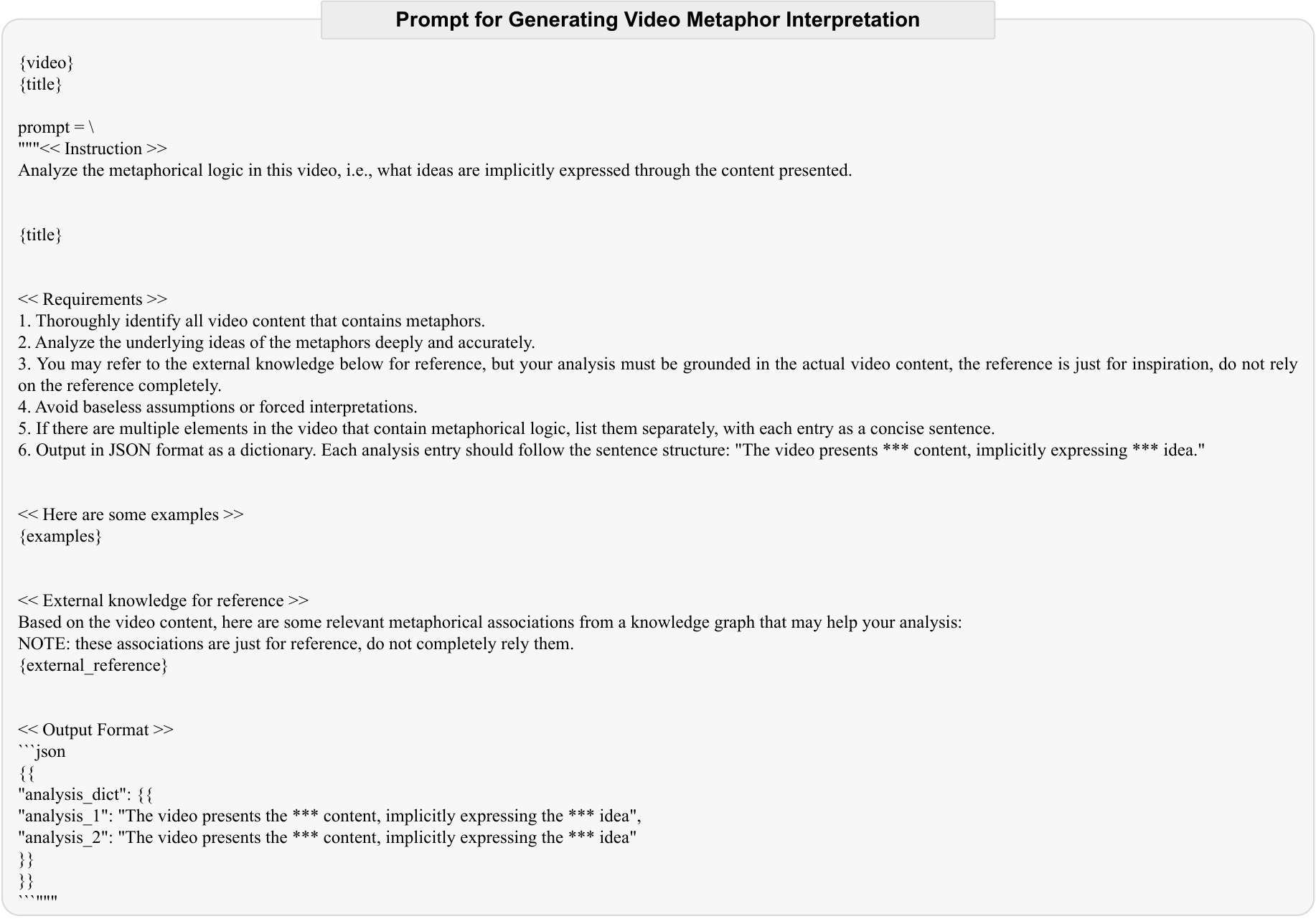} 
\caption{Prompt for generating video metaphor interpretation.}
\label{fig:prompt_for_gene}
\end{figure}

%% file: figs/more_case_1.tex
\begin{figure}[t!]
\centering
\includegraphics[width=\linewidth]{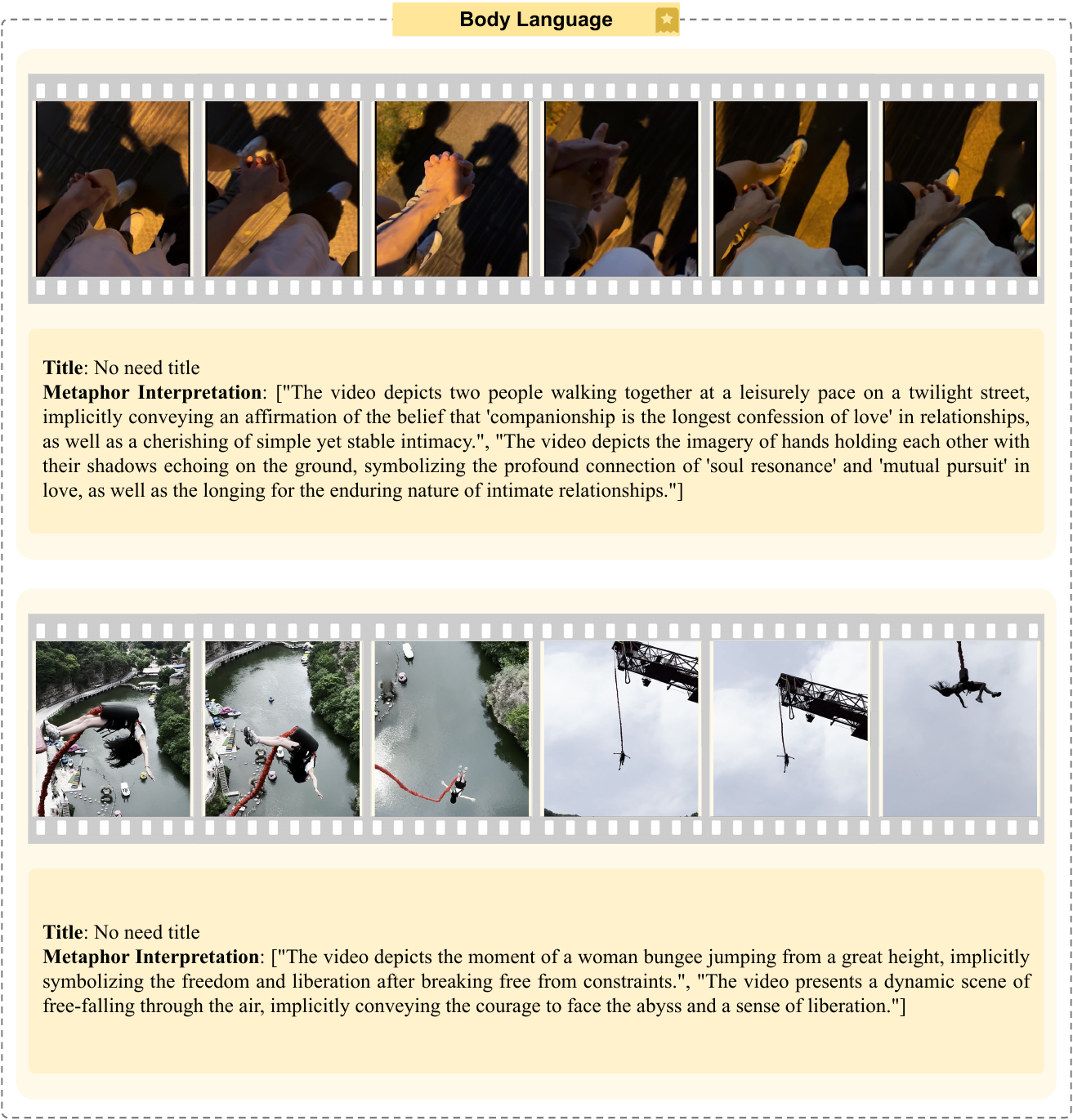} 
\caption{Examples of Body Language. Note that most videos simultaneously contain multiple types of metaphor, we only show the dominant one in each case for convenient illustration.}
\label{fig:more_case_1_fig}
\end{figure}

%% file: figs/more_case_2.tex
\begin{figure}[t!]
\centering
\includegraphics[width=\linewidth]{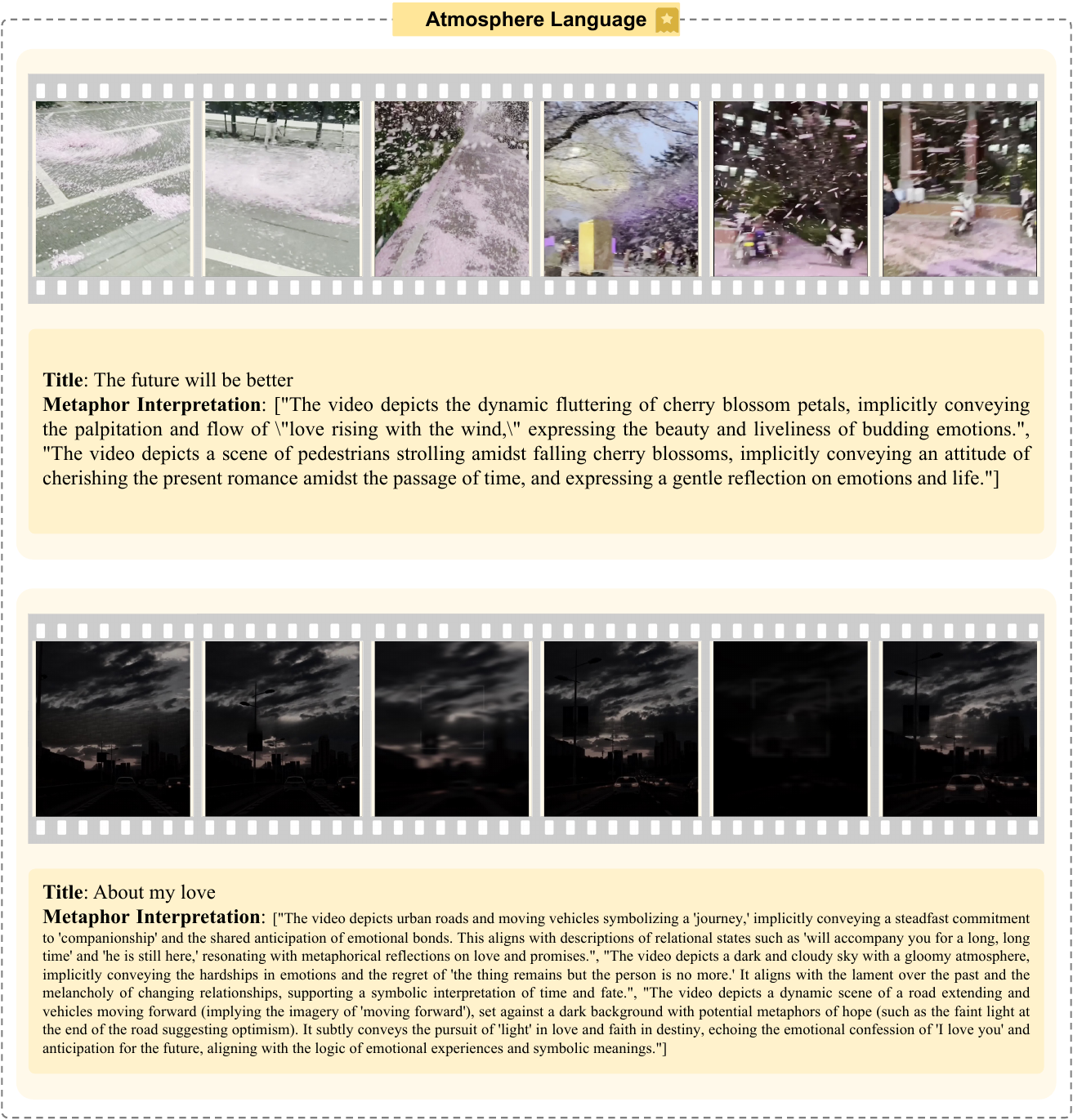} 
\caption{Examples of Atmosphere Language. Note that most videos simultaneously contain multiple types of metaphor, we only show the dominant one in each case for convenient illustration.}
\label{fig:more_case_2_fig}
\end{figure}

%% file: figs/more_case_3.tex
\begin{figure}[t!]
\centering
\includegraphics[width=\linewidth]{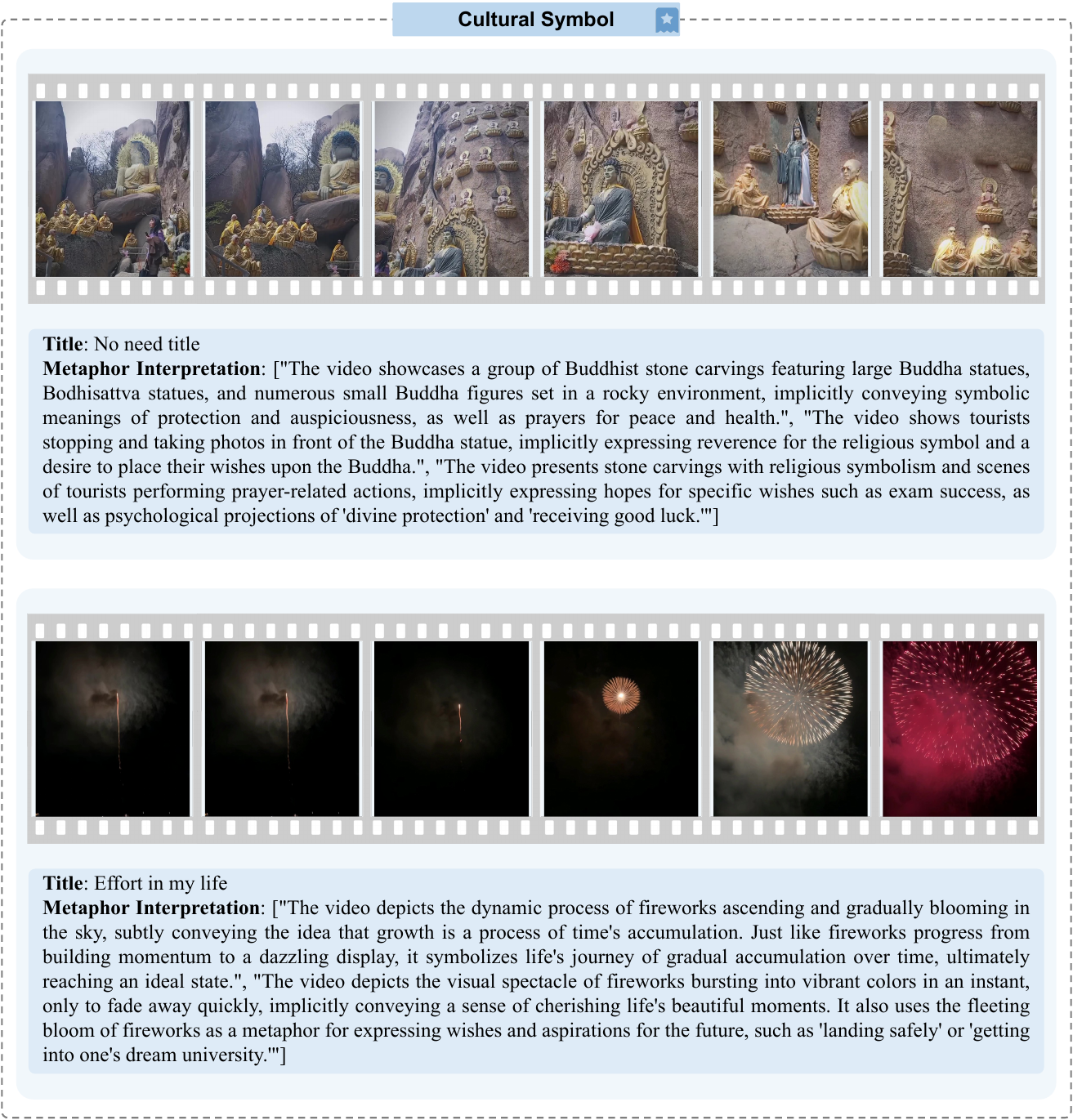} 
\caption{Examples of Cultural Symbol. Note that most videos simultaneously contain multiple types of metaphor, we only show the dominant one in each case for convenient illustration.}
\label{fig:more_case_3_fig}
\end{figure}

%% file: figs/more_case_4.tex
\begin{figure}[t!]
\centering
\includegraphics[width=\linewidth]{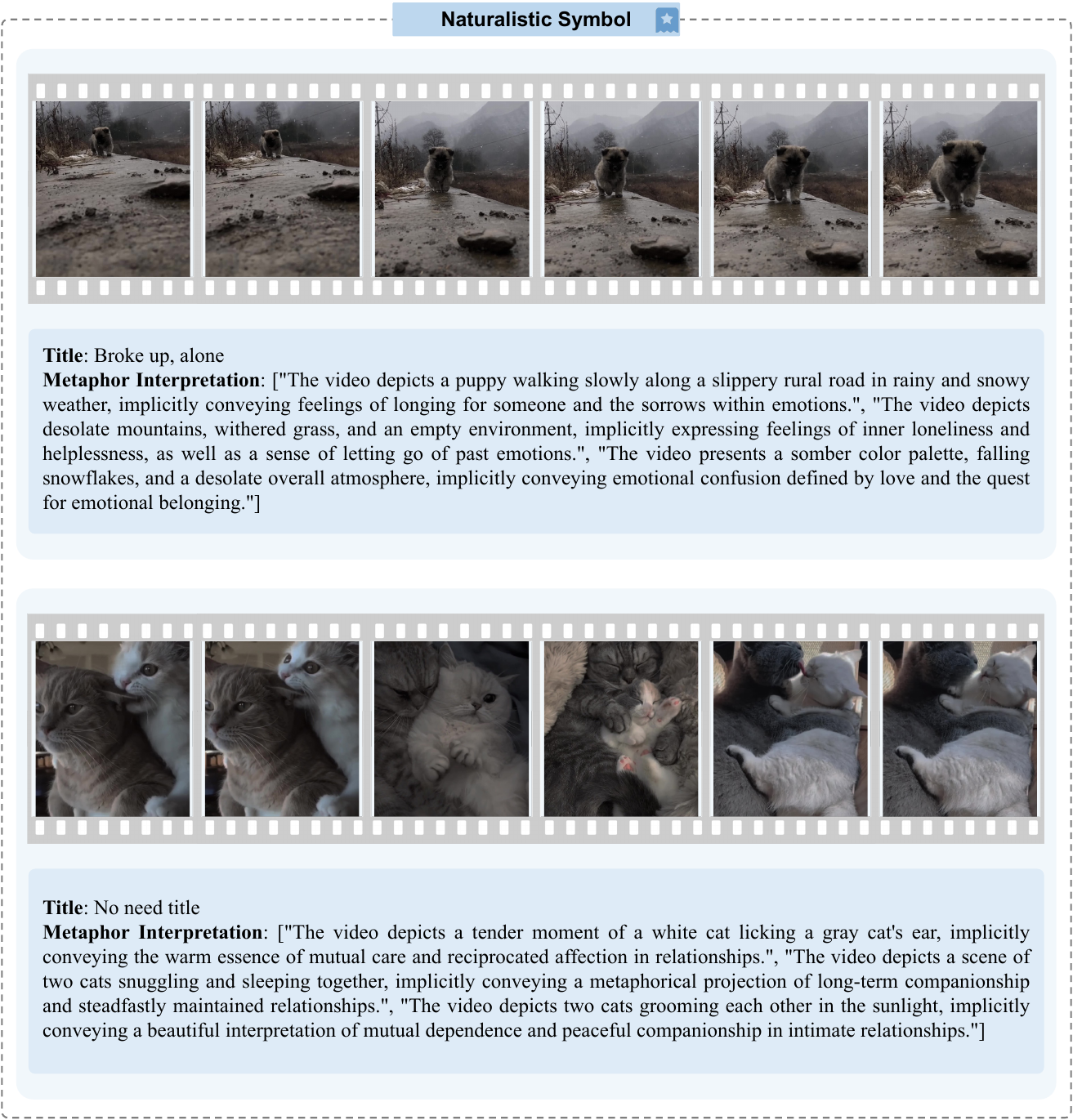} 
\caption{Examples of Naturalistic Symbol. Note that most videos simultaneously contain multiple types of metaphor, we only show the dominant one in each case for convenient illustration.}
\label{fig:more_case_4_fig}
\end{figure}

%% file: figs/more_case_5.tex
\begin{figure}[t!]
\centering
\includegraphics[width=\linewidth]{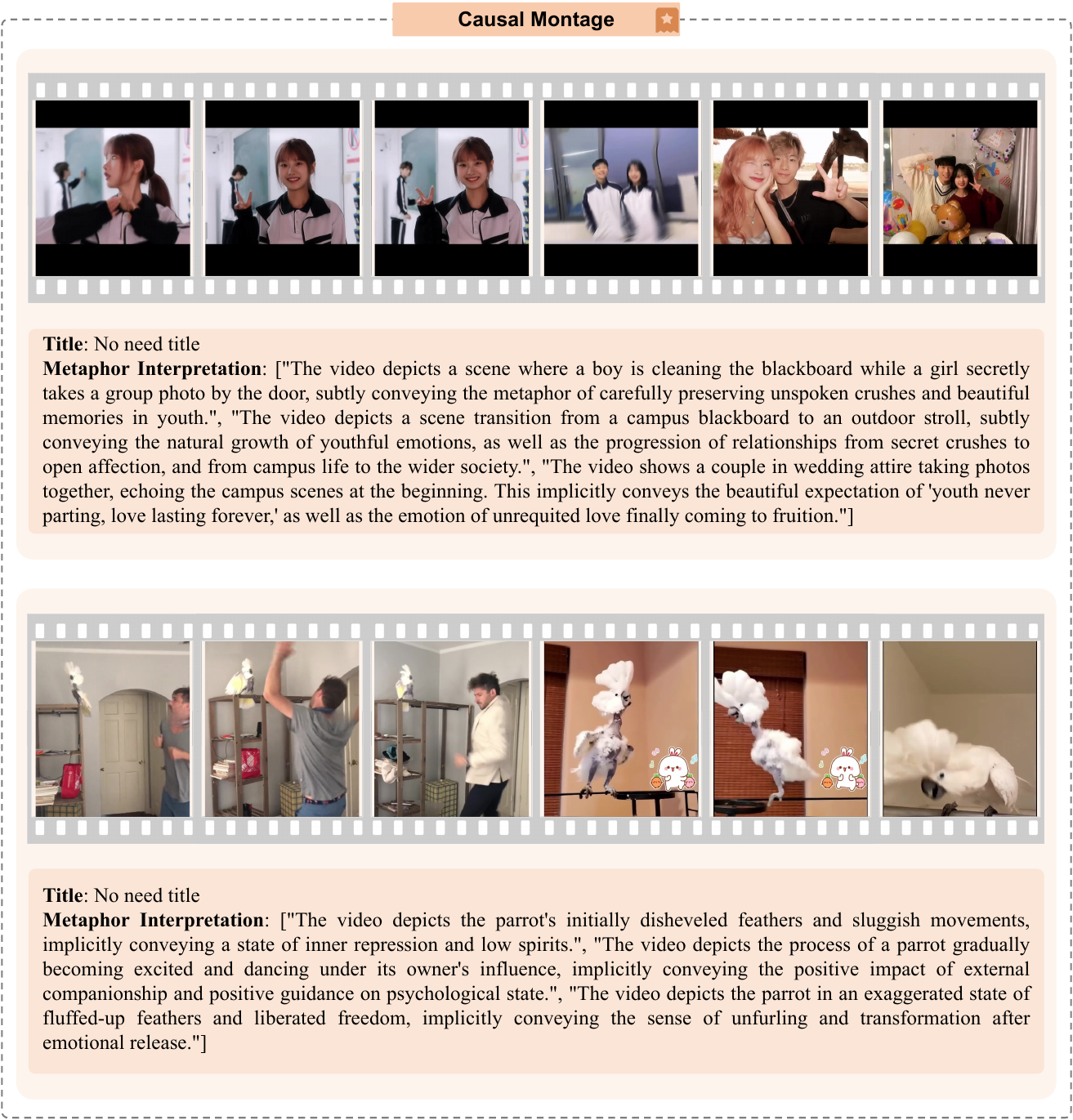} 
\caption{Examples of Causal Montage. Note that most videos simultaneously contain multiple types of metaphor, we only show the dominant one in each case for convenient illustration.}
\label{fig:more_case_5_fig}
\end{figure}

%% file: figs/more_case_6.tex
\begin{figure}[t!]
\centering
\includegraphics[width=\linewidth]{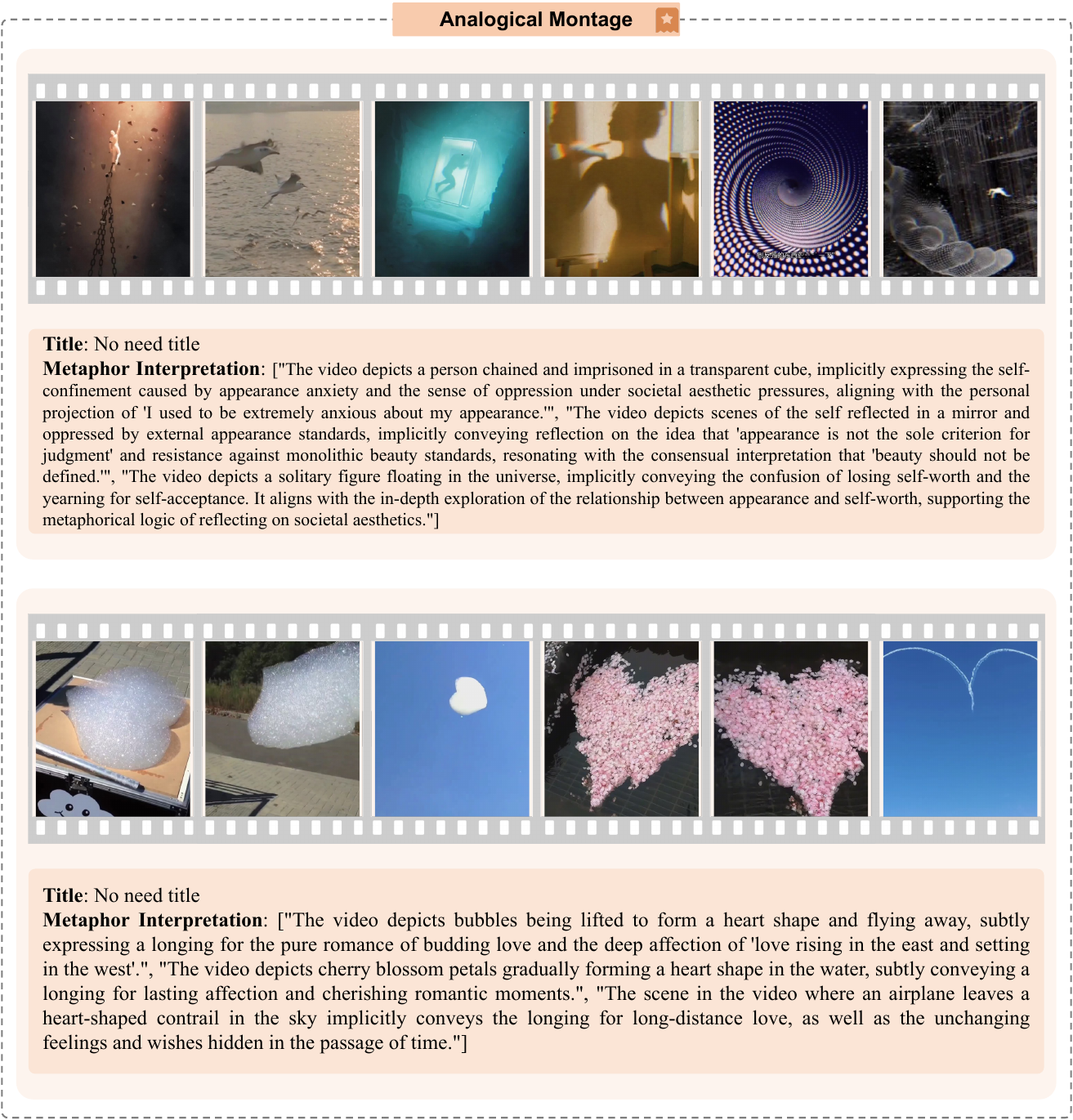} 
\caption{Examples of Analogical Montage. Note that most videos simultaneously contain multiple types of metaphor, we only show the dominant one in each case for convenient illustration.}
\label{fig:more_case_6_fig}
\end{figure}

%% file: figs/more_case_7.tex
\begin{figure}[t!]
\centering
\includegraphics[width=\linewidth]{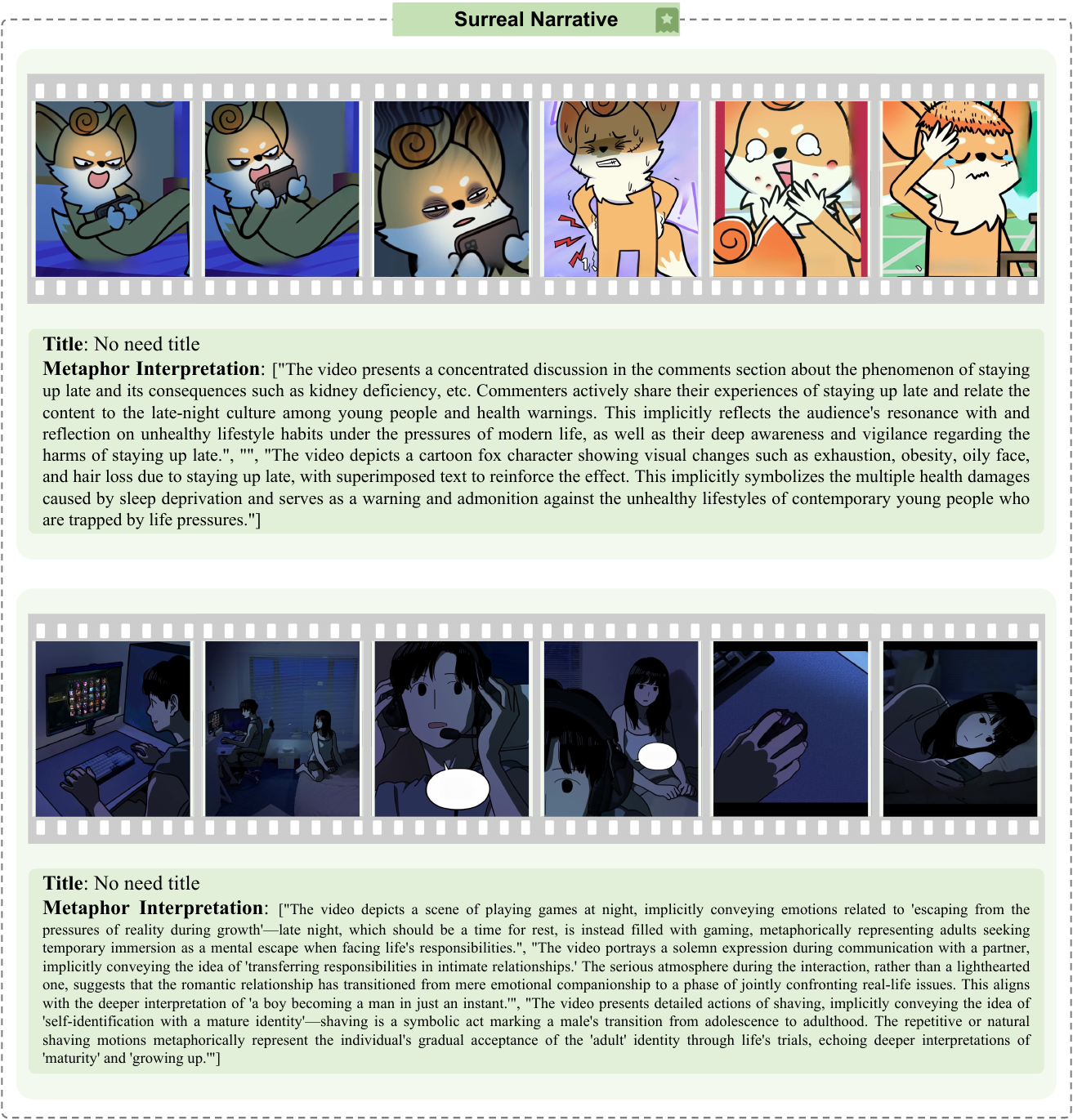} 
\caption{Examples of Surreal Narrative. Note that most videos simultaneously contain multiple types of metaphor, we only show the dominant one in each case for convenient illustration.}
\label{fig:more_case_7_fig}
\end{figure}

%% file: figs/more_case_8.tex
\begin{figure}[t!]
\centering
\includegraphics[width=\linewidth]{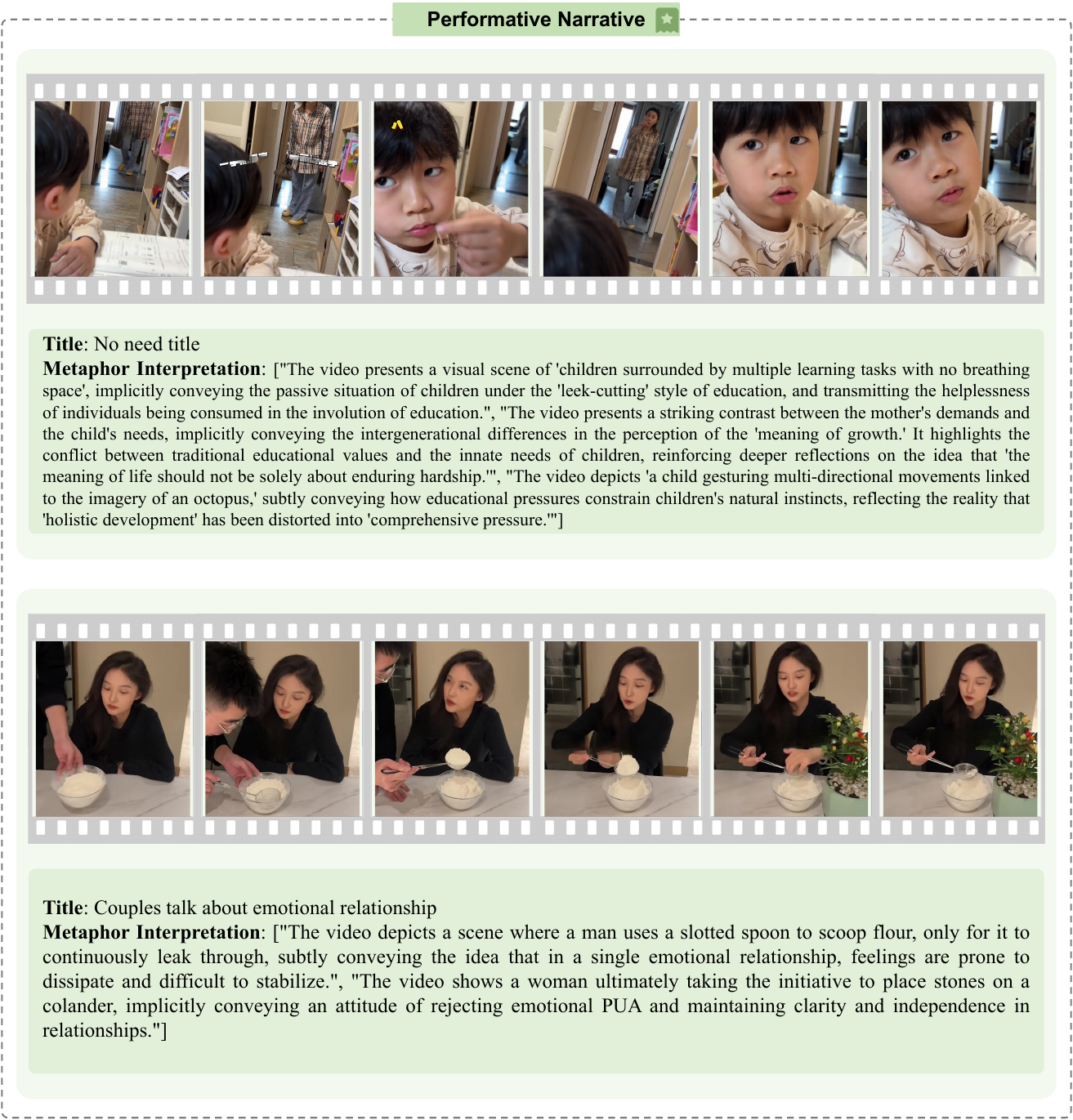} 
\caption{Examples of Performative Narrative. Note that most videos simultaneously contain multiple types of metaphor, we only show the dominant one in each case for convenient illustration.}
\label{fig:more_case_8_fig}
\end{figure}